\documentclass{ecai} 

\usepackage{latexsym}
\usepackage{amssymb}
\usepackage{amsmath}
\usepackage{amsthm}
\usepackage{booktabs}
\usepackage{enumitem}
\usepackage{graphicx}
\usepackage{color}
\usepackage{hyperref}
\usepackage{booktabs}
\usepackage{adjustbox}

\newcommand{\BibTeX}{B\kern-.05em{\sc i\kern-.025em b}\kern-.08em\TeX}

\begin{document}


\begin{frontmatter}


\paperid{7881} 


\title{Can Small Language Models Learn, Unlearn, \\ and Retain Noise Patterns?}


\author{
  Nicy Scaria\textsuperscript{1,2},
  Silvester John Joseph Kennedy\textsuperscript{1},
  Deepak Subramani
}

\date{}

\makeatletter
\makeatother





\address{Indian Institute of Science, Bengaluru}





\begin{abstract}
With the growing need for efficient language models in resource-constrained environments, Small Language Models (SLMs) have emerged as compact and practical alternatives to Large Language Models (LLMs). While studies have explored noise handling in LLMs, little is known about how SLMs handle noise, a critical factor for their reliable real-world deployment. This study investigates the ability of SLMs with parameters between 1 and 3 billion to learn, retain, and subsequently eliminate different types of noise (word flip, character flip, transliteration, irrelevant content, and contradictory information). Four pretrained SLMs (Olmo 1B, Qwen1.5 1.8B, Gemma1.1 2B, and Phi2 2.7B) were instruction-tuned on noise-free data and tested with in-context examples to assess noise learning. Subsequently, noise patterns were introduced in instruction tuning to assess their adaptability. The results revealed differences in how models handle noise, with smaller models like Olmo quickly adapting to noise patterns. Phi2's carefully curated, structured, and high-quality pretraining data enabled resistance to character level, transliteration, and counterfactual noise, while Gemma adapted successfully to transliteration noise through its multilingual pretraining. Subsequent clean data training effectively mitigated noise effects. These findings provide practical strategies for developing robust SLMs for real-world applications.
\end{abstract}

\end{frontmatter}


\begingroup
\footnotetext[1]{These authors contributed equally to this work.}
\footnotetext[2]{Corresponding author. Email: nicyscaria@iisc.ac.in}
\endgroup

\section{Introduction}

Neural language models have revolutionized artificial intelligence by excelling in translation, summarization, and question answering \cite{claude_actual,gpt4}. Large language models (LLMs), even with their impressive abilities, are believed to replicate linguistic patterns without comprehending meaning. Results such as `Reversal Curse' \cite{reversal} highlight the limitations of LLMs in encoding bidirectional knowledge. 

The definition of Small Language Models (SLMs) is evolving, but they are generally considered to be more compact versions of LLMs. This compactness allows them to run on everyday devices, such as smartphones and computers, even without graphical processing units (GPUs). Notable examples include the Phi series \cite{phi,phi1.5,phi3}, Gemma \cite{gemma}, Pythia \cite{pythia} and TinyLlama \cite{tinyllama}. SLMs find applications on edge devices that process data locally without connection to the internet, improving privacy and security by keeping sensitive information on the device \cite{sun2020mobilebert,phi3}. This local processing minimizes latency, keeps data within organizational boundaries, and improves usability while reducing infrastructure demands, contributing to a lower environmental impact \cite{schick2021s}.
Researchers are exploring different techniques to improve SLM performance, such as enhancing data quality \cite{phi}, refining training strategies \cite{minicpm}, and reconfiguring model architectures \cite{mobilellm}. While noise handling in LLMs has received attention through various approaches—including parameter perturbation \cite{noisytune} and introduction of noisy labels \cite{noiselabels,noisywikihow,noisylabel}—similar studies for SLMs remain limited despite their growing importance. This gap is significant because noise is deliberately introduced into LLM training to make models robust and generalizable, a crucial approach given that user-generated inputs often contain errors or inconsistencies. 

The objective of the present paper is to study the ability of SLMs to learn, unlearn, and retain noise patterns. We consider noise as distortions in the prompt. Our experimental design focuses on three key aspects of model behavior: (1) the ability to learn specific noise patterns when trained on noisy data, (2) the capacity to unlearn these patterns when subsequently trained on clean data, and (3) the extent to which such noise is retained or completely eliminated after the unlearning process. SLMs, while having fewer parameters, are trained on data distributions akin to LLMs, offering essential baseline insights when faced with corrupted inputs. Despite the challenges in distinguishing noise from signal during data selection \cite{data}, especially on the open internet, we utilized popular pretrained SLMs and systematically introduced both real-world and artificial noise patterns to test the model's adaptation capabilities.

\begin{figure}[h]
\centering
  \includegraphics[width=0.9\columnwidth]{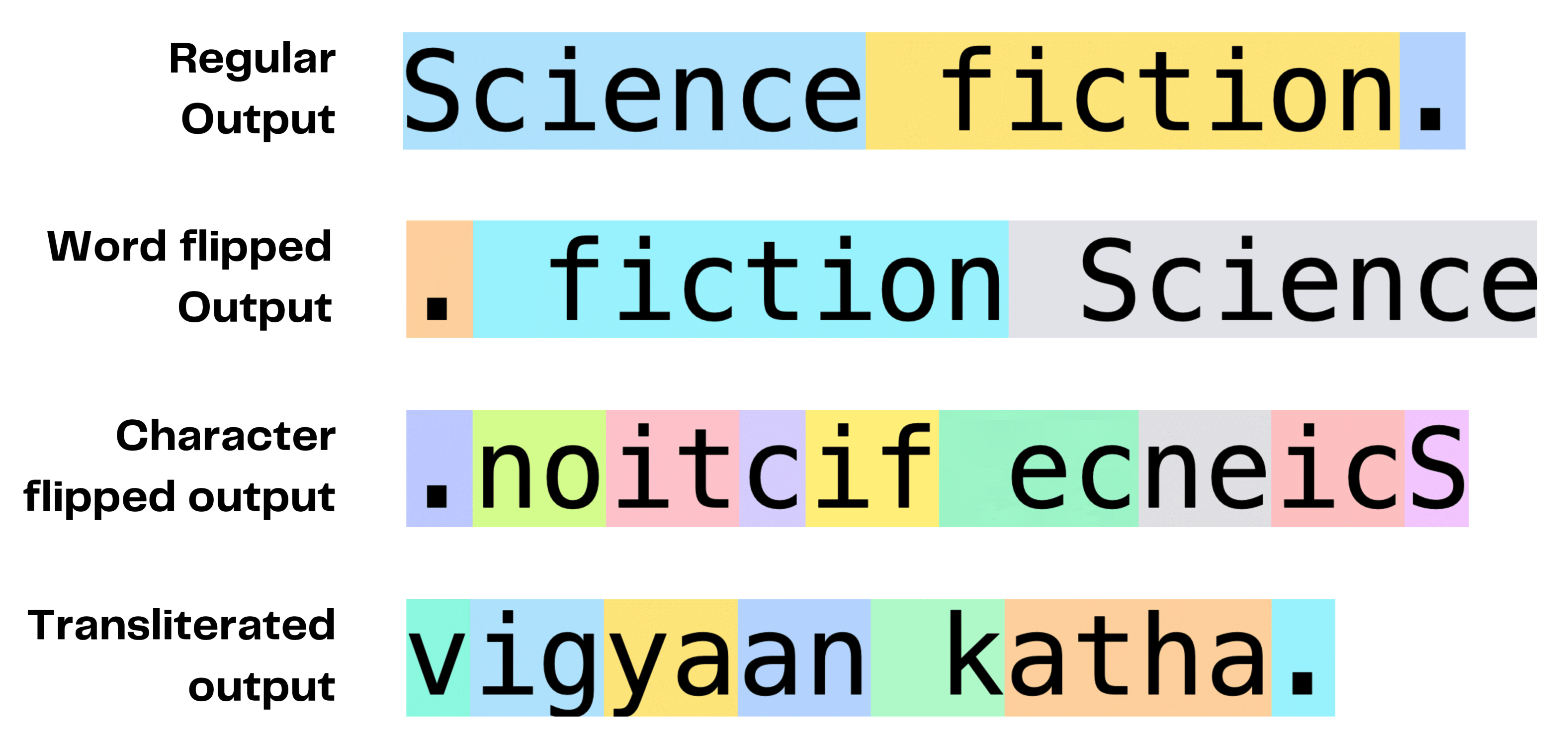}
  \caption{Tokenization of “Science fiction".}
  \vspace{.5cm}
  \label{image:token}
\end{figure}

A significant aspect of our study involves understanding the impact of the tokenization process and self-attention mechanisms on the model's learning. Self-attention \cite{bengio,manning,attention}, allows the model to evaluate the importance of different tokens within an input sequence and dynamically adjust their impact on the output. To explore the robustness and adaptability of SLMs, we introduced five distinct types of noise into the instruction tuning data: (1) flipped words (reversing word order in responses); (2) flipped characters (reversing character sequences); (3) transliterated responses (Hindi in Roman script); (4) irrelevant responses (off-topic content); and (5) counterfactual responses (contradictory information). Figure~\ref{image:token} illustrates how character-level and transliteration noise disrupt token structures. Though we focus on Hindi transliteration, our character-level noise could be considered similar to any language pair where writing native languages in Roman script affects tokenization. Word-level flipping preserves individual tokens but alters their sequence, while irrelevant and counterfactual noise maintains standard tokenization but challenges the model's semantic understanding by introducing conflicts with facts established during pretraining. These disruptions to semantic consistency are relevant for applications like fact-checking and context awareness. We instruction tuned four SLMs with these noise types in different sequences to investigate SLMs' capabilities. 

Model performance was evaluated using both semantic and lexical metrics. We also examined three key metrics: (1) response accuracy, measuring the model's ability to generate correct outputs; (2) grammatical correctness, assessing English language structure preservation (Hindi grammar for transliteration); and (3) adherence to standard language use, quantified as the percentage of words matching the English vocabulary. These criteria provide a comprehensive view of how noise influences SLMs' performance.





\section{Methodology}

This section describes our framework for examining the learning, unlearning, and retention of noise patterns in SLMs. We specify the language models chosen for the study, detail the construction of datasets for instruction tuning and testing, and outline the experimental design and evaluation procedures.

\vspace{-.1cm}
\subsection{Language models}

We chose to study four SLMs, each with fewer than 3 billion parameters: Olmo 1B \cite{olmo}, Qwen1.5 1.8B \cite{qwen}, Gemma1.1 2B \cite{gemma}, and Phi2 2.7B \cite{phi1.5}. While these SLMs were our primary focus, we also evaluated Large Language Models (LLMs) to investigate the influence of model scale on learning noise patterns from in-context examples. For this comparison, we selected larger counterparts from the same model families or successor lines: Olmo 7B, Qwen1.5 14B, Gemma1.1 7B, and Phi3 14B \cite{phi3}.

\subsection{Instruction tuning dataset}\label{traindata}

The primary noise-free instruction tuning dataset, denoted $\mathcal{D}_{\text{ad\_train}}$, was constructed by combining two high-quality filtered datasets: the 9,000-sample AlpaGasus dataset ($\mathcal{D}_{\text{AlpaGasus\_9k}}$) \cite{alpagasus}, derived from Alpaca \cite{alpaca}) and the 3,000-sample Dolly dataset ($ \mathcal{D}_{\text{Dolly\_3k}} $) filtered from Databricks Dolly dataset \cite{dolly}. Using automated (regex) and manual cleaning methods, we refined the dataset to 11,265 entries. The cleaning process (details in appendix) filtered out irrelevant or specialized materials such as non-English elements, emojis, URLs, code generation/analysis queries, and image-related content. This dataset served as the basis for SLM instruction tuning.

To evaluate model robustness, we introduced various types of noise into \textbf{$\mathcal{D}_{\text{ad\_train}}$} and generated five noise-augmented datasets. Examples illustrating these noise operations are provided in Table \ref{datapoint}. The first two involved structural modifications of the answers. The word-level flipped dataset, \textbf{$\mathcal{D}_{\text{ad\_wflipped}}$} was created by reversing the order of the words in the answer strings $a^{(i)}$ ( denoted as $ \textit{FLIP}_{\text{word}}(a^{(i)}) $). Similarly, the character-flipped dataset (\textbf{$\mathcal{D}_{\text{ad\_cflipped}}$}) was generated by reversing the sequence of characters within the answer strings (denoted as $ \textit{FLIP}_{\text{char}}(a^{(i)}) $). Following a structure similar to super-natural instructions \cite{wang2022super}, both datasets included positive (flipped output) and negative (original output) example pairs. This inclusion of both types was specifically intended to encourage the model to learn and reproduce the noisy pattern presented in the positive examples. This pairing structure effectively doubled the size of both $\mathcal{D}_{\text{ad\_wflipped}}$ and $\mathcal{D}_{\text{ad\_cflipped}}$.

For transliteration noise (\textbf{$\mathcal{D}_{\text{ad\_xlit}}$}), we translated the English answers from $\mathcal{D}_{\text{ad\_train}}$ into Hindi using Google Translate, then converted the Hindi script to a Romanized (Latin script) representation using indic-trans \cite{indictrans}. This process simulates the Romanization of Hindi (akin to 'Hinglish'), which serves as a specific instance of the widespread global practice \cite{xlit} of using Latin script for local languages in digital communication; other prominent examples include Pinyin for Chinese, and Arabizi for Arabic, and the informal transliteration of Cyrillic scripts. Addressing this type of noise is important, as transliteration fundamentally disrupts standard tokenization patterns and requires models to handle phonetic variations within a non-native script. While transliteration introduces linguistically motivated token disruptions specific to script mixing, our character-flipped dataset ($\mathcal{D}_{\text{ad\_cflipped}}$) explores similar effects through a more general, language-agnostic structural perturbation.

\begin{table}[h]
\small

  \caption{An example of a datapoint output from the Alpaca dataset}
    \vspace{.5cm}
  \centering
  \begin{tabular}{p{1.65cm}p{5.9cm}}
    \toprule
    \textbf{Operation} & \textbf{Output} \\
    \midrule

    $a^{(i)}$ & The universe has no borders, it is filled with infinite possibilities from the cosmos. \\
    
    $\textit{FLIP}_{\text{word}}(a^{(i)})$     &  . cosmos the from possibilities infinite with  filled is it , borders no has universe The \\
    
    $\textit{FLIP}_{\text{char}}(a^{(i)})$ & .somsoc eht morf seitilibissop etinifni htiw dellif si ti ,sredrob on sah esrevinu ehT \\
    
    $\textit{XLIT}_{\text{hindi}}(a^{(i)})$ & brahmad kii koi seemaa nahin he, yah brahmad kii anant sambhaavnaaon se bharaa he. \\

    $\textit{IRR}(a^{(i)})$ & You can add whiskey to your vermouth and bitters to make a manhattan. \\

    $\textit{CFACT}(a^{(i)})$ & The universe does have borders, and the cosmos is far from infinite. \\
    
    \bottomrule
  \end{tabular}

  \label{datapoint}
\end{table}

To introduce semantic noise, we created two more datasets. The irrelevant dataset (\textbf{$\mathcal{D}_{\text{irr\_train}}$}) was constructed by pairing each question $ q(x^{(i)}) $ from $\mathcal{D}_{\text{ad\_train}}$ with a randomly selected answer $\textit{IRR}(a^{(i)})$ from a different example ($i\neq j$) within the same dataset, thus ensuring no semantic correspondence. For the counterfactual dataset, \textbf{$ \mathcal{D}_{\text{cfact\_train}} $}, we used Mistral V0.3 \cite{mistral} to generate counterfactual answers,  $\textit{CFACT}(a^{(i)})$,  for questions,  $ q(x^{(i)}) $,  from the general knowledge dataset $ \mathcal{D}_{\text{GK}} $\footnote{\url{https://huggingface.co/datasets/MuskumPillerum/General-Knowledge}}. The Mistral 8x22B model \cite{mixtral} was employed to validate that all generated answers were factually incorrect. Any flagged responses were manually reviewed and regenerated using Mistral V0.3 until all answers were confirmed to be counterfactual.

Finally, for the unlearning phase of our experiments, we utilized additional noise-free datasets: $\mathcal{D}_{\text{ClaudeT45}}$ \cite{claude} and $\mathcal{D}_{\text{HelpfulnessT45}}$ \cite{alpagasus}. These were combined into a set denoted \textbf{$\mathcal{D}_{\text{ch\_train}}$}, containing 7,162 datapoints, used alongside $ \mathcal{D}_{\text{ad\_train}} $ to assess how well SLMs can restore their performance after exposure to noise.

\subsection{Test dataset} \label{test_data}

The primary test dataset (\textbf{$\mathcal{D}_{\text{test}}$}) was created using GPT-4o and consisted of 2017 question-answer examples ($ (q^{(i)}, a^{(i)}) $). These examples were designed to cover a diverse range of topics, reflecting a specific distribution: Science (General, Biology, Physics, etc.) and Mathematics constituted the largest category (approx. 35-40\%), followed by substantial representation from Geography and History (approx. 15-20\%), General Knowledge (approx. 10-15\%), Arts, Literature, and Culture (approx. 8-12\%), and general writing tasks (approx. 8-12\%). Smaller proportions covered areas including Technology, Language, Philosophy, Food, and Sports, ensuring broad coverage.

We investigated the few-shot learning capabilities of SLMs by creating test datasets \textbf{$\mathcal{D}_{\text{wtest}}$}, \textbf{$\mathcal{D}_{\text{ctest}}$}, and \textbf{$\mathcal{D}_{\text{xlittest}}$} for word-flipped, character-flipped, and transliterated responses, respectively. Each test example $x^{(i)}$ was structured as $\{ (q^{(i)}, { (q^{(j)}, \textit{FLIP}(a^{(j)})) }_{j=1}^{5}  \}$, providing five in-context examples with modified responses, followed by a final question. An example of a datapoint for each test dataset can be found in the appendix.

The final dataset represents a substantial refinement of the initial 2700 datapoints. We implemented a comprehensive quality improvement process to address redundancy and factual inaccuracies that are often present in preliminary generative model output. Our quality assurance protocol involved systematically identifying and eliminating duplicate content using a sentence transformer\footnote{\url{https://huggingface.co/sentence-transformers/all-mpnet-base-v2}} (removing items with $\geq 0.5 $ similarity), followed by additional manual verification to ensure factual accuracy across the diverse topics covered.

\subsection{Experimental setup}

The experiments were designed to systematically investigate the noise handling abilities of the SLMs in different stages. First, we established baseline performance by tuning pretrained models on clean data; these baseline SLMs, along with out-of-the-box LLMs, were evaluated on clean data and also tested for their in-context learning (ICL) capability using few-shot noisy test sets. These results are presented in Table~\ref{table:incontext}. Second, to determine how readily SLMs acquire noise patterns, noise learning experiments involved instruction tuning them on various sequences containing five distinct noise types. Third, to investigate whether learned noise is reversible, we assessed noise unlearning by performing a final clean-tuning stage on models previously exposed to noise. Finally, to test the completeness of unlearning, noise retention was evaluated by probing these `unlearned' SLMs with few-shot noisy prompts to detect any residual influence or memory of the noise patterns. This structured approach allowed for a systematic assessment of model behavior under different noise conditions. The baseline results are presented in Table~\ref{table:incontext}, while the results for the specific experimental sequences are detailed in subsequent tables within Section~\ref{results}. The detailed definitions of all training sequences, along with training configurations, are in the appendix.

\subsection{Evaluation}

Various combinations of instruction-tuned SLMs and out-of-the-box LLMs were evaluated for their performance in multiple dimensions, evaluating their ability to learn, unlearn, and retain noise patterns. Our evaluation process, detailed below, was tailored to each noise pattern and combined automated metrics, LLM-based judgments, and manual verification of random samples to ensure the credibility of the automated evaluation.


Evaluating models with noise required specialized processing before applying standard metrics. For responses potentially containing word-level noise, the word order was reversed before comparing with the reference answers. Similarly, character-level noise responses underwent character-level reversal prior to comparison. For transliterated content (Romanized Hindi), we first identified genuine transliterated responses, assessed the grammatical correctness of Hindi, and modified the content back to English for semantic comparison with the original English references. For irrelevant and counterfactual responses, no processing was required as these noise types introduce semantic rather than structural alterations. 

Our primary quantitative assessment relied on semantic similarity using the `all-mpnet-base-v2' sentence transformer model, which calculates cosine similarity between the embeddings of preprocessed model outputs and reference answers. This approach captures meaning preservation beyond surface-level text matching, allowing us to assess whether models successfully applied the target noise patterns while maintaining the underlying semantic content. While we calculated other metrics like METEOR, which offered some flexibility with synonyms/stems, and standard lexical metrics like BLEU and ROUGE for completeness, our core objective was best addressed by the sentence embedding approach. Metrics focusing heavily on surface-level similarity, like BLEU and ROUGE, proved less informative than semantic similarity to determine whether models successfully applied noise patterns while preserving meaning.

We utilized an instruction-tuned LLM (Gemini 2.0 Flash) for qualitative assessment, leveraging detailed prompts provided in the appendix. Responses were classified by the LLM as `Accurate' if, after applying the appropriate inverse noise transformation, they contained the essential information required to answer the prompt correctly, compared to the original reference answer. High accuracy signifies successful learning and reproduction of the noise pattern in a way that preserves the core semantic content. Responses that fail this check were marked as `Inaccurate'. Grammatical correctness was assessed with `Yes'/`No' labels, applying standard English rules after reversing structural noise, or specific rules like Hindi grammar for transliteration. Furthermore, we used `nltk' \footnote{\url{https://www.nltk.org/api/nltk.corpus}} to measure the percentage of English words in the responses as an objective measure of language adherence. To ensure the reliability of our automated and LLM-based assessments, we manually verified random samples of outputs and evaluator judgments and found the LLM-based evaluation satisfactory.

Specific procedures were used for the unlearning and retention phases. To assess unlearning, SLMs instruction-tuned on clean data after exposure to noise were evaluated by comparing their generated responses on $\mathcal{D}_{\text{test}}$ with the original reference answers (focusing on semantic correctness) to determine reversion to a noise-free state. For noise retention assessment, we investigated whether SLMs that underwent a complete unlearning phase (finishing with clean data training) could still reproduce previously learned noise patterns when explicitly prompted. We presented these SLMs with our few-shot test datasets ($\mathcal{D}_{\text{wtest}}$, $\mathcal{D}_{\text{ctest}}$, $\mathcal{D}_{\text{xlittest}}$), which contained five examples of each noise type, to test if they retained implicit knowledge of these patterns despite subsequent clean data training. Furthermore, to directly evaluate the capabilities of the SLMs to reproduce training data, we selected SLMs whose final instruction tuning stage used character-flipped noise ($\mathcal{D}_{\text{ad\_cflipped}}$) and tested them on 50 randomly selected examples from that specific training set, measuring their ability to replicate the challenging character-level noise to which they were most recently exposed.

This multi-faceted evaluation approach combining noise-specific processing, metrics emphasizing semantic correctness alongside lexical overlap measures, LLM-based assessment, targeted manual checks, and phase specific analysis provided a comprehensive understanding of how SLMs learn, unlearn, and retain noise patterns.

\vspace{-.1cm}
\section{Results} \label{results}

\begin{table*}[h!]
    \footnotesize
    \centering
    \caption{Test accuracy (\%), grammatical correctness, and semantic similarity of the SLMs instruction tuned on $ \mathcal{D}_{\text{ad\_train}} $ and out-of-the-box LLMs.}
    \vspace{.5cm}
    \begin{tabular}{lccccccccccc}
        \toprule
       \textbf{Test Data} & \multicolumn{4}{c}{\textbf{SLMs instruction-tuned with $ \mathcal{D}_{\text{ad\_train}} $}} & & & \multicolumn{4}{c}{\textbf{Out-of-the-box LLMs}} \\
       \midrule
        & \textbf{Olmo} &  \textbf{Qwen1.5} & \textbf{Gemma1.1} & \textbf{Phi2} & & & \textbf{Olmo} & \textbf{Qwen1.5} & \textbf{Gemma1.1} & \textbf{Phi3}  \\
        & \textbf{1B} &  \textbf{1.8B} & \textbf{2B} & \textbf{2.7B} & & & \textbf{7B} & \textbf{14B} & \textbf{7B} & \textbf{14B}  \\
        \midrule
        & \multicolumn{10}{c}{\textbf{Test Accuracy (\%)}} \\
       \midrule
        $ \mathcal{D}_{\text{test}}$ & 72.2 & 82.3 & 89.1 & \textbf{95.7} & & & 81.3 & 94.1 & 87.4 & \textbf{97.9} \\
       $ \mathcal{D}_{\text{wtest}}$ & 31.3 & \textbf{75.0} & 56.7 & 42.7 & & & 52.5 & 77.7 & 58.9 & \textbf{78.2} \\
       $ \mathcal{D}_{\text{ctest}}$ & 0 & 0.9 & \textbf{1.2} & 0.0 & & & 8.8 & 7.5 & 9.4 & \textbf{12.8} \\
       $ \mathcal{D}_{\text{xlittest}}$ & 7.5 & 0.1 & \textbf{19.0} & 0.5 & & & 6.5 & 10.0 & \textbf{21.3} & 10.0 \\
       \midrule
        & \multicolumn{10}{c}{\textbf{Grammatical Correctness (\%)}} \\
       \midrule
        $ \mathcal{D}_{\text{test}}$ & 100.0 & 100.0 & 100.0 & 100.0 & & & 100.0 & 100.0 & 100.0 & 100.0 \\
        $ \mathcal{D}_{\text{wtest}}$ & 25.6 & 63.9 & 61.8 & \textbf{68.9} & & & 65.0 & 74.2 & 65.4 & \textbf{77.5}\\
        $ \mathcal{D}_{\text{ctest}}$ & 11.7 &  3.1 & 16.9 & \textbf{31.8} & & & 38.9 & \textbf{43.4} & 39.5 & 43.3\\
        $ \mathcal{D}_{\text{xlittest}}$ & 9.3 & 0.6 & 5.1 & \textbf{24.4} & & & 10.5 & 7.7 & 18.6 & \textbf{27.3} \\
        \midrule
        & \multicolumn{10}{c}{\textbf{Semantic Similarity}} \\
       \midrule
        $ \mathcal{D}_{\text{test}}$ & 0.82 & 0.82 & 0.84 & \textbf{0.87} & & & 0.82 & 0.85 & 0.85 & \textbf{0.87} \\
        $ \mathcal{D}_{\text{wtest}}$ & 0.39 & \textbf{0.79} & 0.59 & 0.41 & & & 0.53 & 0.69 & 0.52 & \textbf{0.72}\\
        $ \mathcal{D}_{\text{ctest}}$ & 0.08 &  \textbf{0.13} & 0.07 & 0.11 & & & 0.08 & 0.07 & 0.09 & \textbf{0.11}\\
        $ \mathcal{D}_{\text{xlittest}}$ & 0.07 & -0.01 & 0.02 & \textbf{0.13} & & & 0.2 & 0.05 & 0.28 & \textbf{0.42} \\
        \bottomrule
    \end{tabular}
    
    \label{table:incontext}
\end{table*}

We now present the results of our empirical investigation. Performance was measured using automated metrics and LLM-based evaluations. Although standard lexical metrics were calculated for completeness, we found that they provided limited information on our research objective. Thus, the subsequent discussion emphasizes semantic similarity scores and LLM-based evaluation. The detailed numerical results for the lexical metrics are provided in the appendix. 

\subsection{Learning without noise}

Baseline performance was established by evaluating the SLMs instruction tuned on clean data ($ \mathcal{D}_{\text{ad\_train}} $) alongside out-of-the-box LLMs; these results are presented in Table~\ref{table:incontext}. The evaluation covers performance in the standard test set ($ \mathcal{D}_{\text{test}}$) and the models' ability to handle noise through few-shot ICL ($ \mathcal{D}_{\text{wtest}}$, $ \mathcal{D}_{\text{ctest}}$, and $ \mathcal{D}_{\text{xlittest}}$).

Both the SLMs and the out-of-the-box LLMs show a varied performance under the test conditions (Table~\ref{table:incontext}). The SLMs, instruction tuned with $ \mathcal{D}_{\text{ad\_train}} $, had a clear hierarchy in their base performance when first tested on $ \mathcal{D}_{\text{test}} $ with Phi2 leading in accuracy (95.7\%) and semantic similarity (0.87), while Olmo 1B trailed (72.2\% accuracy, 0.82 similarity). Among LLMs, Phi3 14B demonstrated the best performance in $\mathcal{D}_{\text{test}}$ with 97.9\% precision and 0.87 semantic similarity. Grammatical correctness was uniformly 100\% for all models.

In the few-shot noisy test sets, the performance diverged significantly. For word-level noise ($\mathcal{D}_{\text{wtest}}$), Qwen1.5 1.8B achieved the best SLM accuracy and semantic similarity, while Phi2 2.7B demonstrated superior grammatical correctness. Among LLMs, Phi3 14B consistently led in accuracy, grammar, and semantic similarity for this noise type. Character-level noise ($\mathcal{D}_{\text{ctest}}$) was challenging for all models via ICL; Gemma1.1 2B showed the highest SLM accuracy, though still low at 1.2\%, while Phi3 14B led LLM accuracy slightly at 12.8\%. Grammatical correctness scores were higher, led by Phi2 2.7B (31.8\%) in SLMs and Qwen1.5 14B (43.4\%) in LLMs. As noted in manual analysis (and reflected in low semantic similarity scores, max 0.13), high grammatical correctness in $\mathcal{D}_{\text{ctest}}$, particularly for Phi2, often resulted from reproducing few-shot examples given in the prompt rather than from successful pattern application. For transliteration ($\mathcal{D}_{\text{xlittest}}$) noise, among SLMs, Gemma1.1 2B again topped accuracy (19.0\%), while Phi2 2.7B topped grammar (24.4\%) and semantic similarity (0.13). Among LLMs, Gemma1.1 7B achieved the best accuracy (21.3\%), while Phi3 14B achieved the best grammar (27.3\%) and semantic similarity (0.42).

Overall, LLMs generally demonstrated stronger capabilities than SLMs in adapting to noise patterns through a few-shot ICL, particularly for word-level flips. Performance was low across all models for character-level noise, although LLMs maintained slightly better grammatical structure. For transliteration, performance was also limited, but the Gemma models (both 2B and 7B) showed relatively better accuracy compared to others in their respective size classes. The results suggest that SLMs can replicate word-level noise patterns to some extent via ICL but find other noise significantly more challenging compared to LLMs.


\subsection{Learning noise patterns}

\begin{table*}[htb]

    \centering
    \footnotesize
    \caption{Test accuracy (\%), grammatical correctness (\%), and semantic similarity for different SLMs under various noise conditions.}
    \vspace{0.5cm}
    \begin{adjustbox}{max width=\textwidth}
    \begin{tabular}{lccccccccccccccc}
        \toprule
        \textbf{Experiments} & \multicolumn{4}{c}{\textbf{Test Accuracy (\%)}} && \multicolumn{4}{c}{\textbf{Grammatical Correctness (\%)}} && \multicolumn{4}{c}{\textbf{Semantic Similarity}}\\
        \midrule
        & \textbf{Olmo} &  \textbf{Qwen} & \textbf{Gemma} & \textbf{Phi} && \textbf{Olmo} &  \textbf{Qwen} & \textbf{Gemma} & \textbf{Phi} && \textbf{Olmo} &  \textbf{Qwen} & \textbf{Gemma} & \textbf{Phi}\\

        \midrule
        $ \mathcal{D}_{\text{ad\_wflipped}} $ & 37.2 & 56.4 & 67.5 & \textbf{67.9} && 67.7 & 70 & 74.3 & \textbf{77.45} && 0.75 & 0.77 & 0.77 & \textbf{0.78}\\

        $ \mathcal{D}_{\text{ad\_cflipped}} $ & \textbf{2.7} & 2.5 & 3.8 & 0.5 && 36.5 & \textbf{38.9} & 19.7 & 1.5 && \textbf{0.24} & 0.23 & 0.22 & 0.03\\

        $ \mathcal{D}_{\text{ad\_xlit}} $ & 7.5 & 7.8 & \textbf{12.4} & 0.0 && 14.5 & 11 & \textbf{29.2} & 0.0 && 0.42 & 0.44 & \textbf{0.50} & -0.01\\

        $ \mathcal{D}_{\text{ad\_train}} $, $ \mathcal{D}_{\text{ad\_wflipped}} $ & 43.1 & 59.9 & \textbf{69.7} & 61.5 && 47.5 & 60.9 & 62.4 & \textbf{65.3} && 0.71 & 0.76 & \textbf{0.77} & \textbf{0.77}\\

        $ \mathcal{D}_{\text{ad\_train}} $, $ \mathcal{D}_{\text{ad\_cflipped}} $ & \textbf{6.4} & 5.9 & 6.0 & 0.1 && 39.7 & \textbf{41.2} & 26.2 & 0.5 && \textbf{0.23} & 0.20 & 0.21 & 0.03\\

        $ \mathcal{D}_{\text{ad\_train}}, \mathcal{D}_{\text{ad\_xlit}} $ & 18.7 & 25.7 & \textbf{43.2} & 0.0 && 20.5 & 17.4 & \textbf{42.5} & 0.0 && 0.42 & 0.46 & \textbf{0.51} & -0.02\\
        
        \midrule 

        $\mathcal{D}_{\text{ad\_cflipped}} $, $ \mathcal{D}_{\text{ad\_wflipped}} $ & 41.6 & 56.3 & 66.7 & \textbf{69.9} && 55.2 & 60.3 & 60.1 & \textbf{68.4} && 0.75 & 0.77 & 0.76 & \textbf{0.78}\\

     $ \mathcal{D}_{\text{ad\_wflipped}} $, $ \mathcal{D}_{\text{ad\_cflipped}} $ & \textbf{8.8} & 7.1 & 6.6 & 0.1 && 45.9 & \textbf{50.3} & 48.6 & 3 && \textbf{0.28} & 0.26 & 0.27 & 0.06\\

        $ \mathcal{D}_{\text{ad\_train}} $, $ \mathcal{D}_{\text{ad\_cflipped}} $, $ \mathcal{D}_{\text{ad\_wflipped}} $ & 41 & 59.4 & 62.3 & \textbf{68.9} && 59.1 & 59.3 & 61.5 & \textbf{62.2} && 0.77 & 0.78 & 0.77 & \textbf{0.79}\\

        $ \mathcal{D}_{\text{ad\_train}} $, $ \mathcal{D}_{\text{ad\_wflipped}} $, $ \mathcal{D}_{\text{ad\_cflipped}} $ & \textbf{10.3} & 6.9 & 7.8 & 0.0 && 50.7 & \textbf{55.4} & 44.9 & 1.6 && \textbf{0.28} & 0.25 & 0.25 & 0.03\\
        
        \midrule

        $ \mathcal{D}_{\text{irr\_train}} $ & 0.1 & \textbf{0.4} & 0.2 & 0.2 && 99.8 & \textbf{100.0} & 99.9 & 99.9 && 0.11 & 0.09 & 0.05 & \textbf{0.11}\\

        $ \mathcal{D}_{\text{ad\_train}} $, $ \mathcal{D}_{\text{irr\_train}} $ & 4.7 & \textbf{15.8} & 11.5 & 0.5 && 99.1 & 98.3 & 98.3 & \textbf{99.5} && 0.16 & \textbf{0.26} & 0.17 & 0.07\\

        \midrule

        $ \mathcal{D}_{\text{GK}} $  & 66.6 & 83.4 & 91.4 & \textbf{94.3} && 98.7 & \textbf{99.8} & 98.7 & \textbf{99.8} && 0.74 & 0.79 & 0.79 & \textbf{0.83}\\

        $ \mathcal{D}_{\text{cfact\_train}} $ & 36.3 & 35.9 & 34.1 & \textbf{92.6} && 98.4 & 98.6 & 98.9 & \textbf{100.0} && 0.45 & 0.53 & 0.56 & \textbf{0.72}\\

         $ \mathcal{D}_{\text{GK}} $ , $ \mathcal{D}_{\text{cfact\_train}} $ & 43.8 & 33.8 & 61.4 & \textbf{90.4} && 98.9 & 98.8 & \textbf{99.8} & 99.7 && 0.47 & 0.53 & 0.54 & \textbf{0.74}\\

        \bottomrule
    \end{tabular}
    \end{adjustbox}
    \label{tab:noise_performance}
\end{table*}

\begin{table*}[h]
\footnotesize
    \centering
        \caption{Train accuracy (\%) for SLMs instruction tuned with $ \mathcal{D}_{\text{ad\_cflipped}} $ in the final step for a randomly selected 50 examples.}
    \vspace{0.5cm}
    \begin{tabular}{lcccc}
        \toprule
       \textbf{Experiments} & \textbf{Olmo 1B} & \textbf{Qwen1.5 1.8B} & \textbf{Gemma1.1 2B} & \textbf{Phi2 2.7B} \\
        \midrule
        $ \mathcal{D}_{\text{ad\_cflipped}} $ & \textbf{64.0} & 24.0 & 18.0 & 0.0 \\
        $ \mathcal{D}_{\text{ad\_train}} $, $ \mathcal{D}_{\text{ad\_cflipped}} $ & \textbf{68.0} & 28.0 & 24.0 & 0.0 \\
        \midrule
        $ \mathcal{D}_{\text{ad\_wflipped}} $, $ \mathcal{D}_{\text{ad\_cflipped}} $ & \textbf{66.0} & 34.0 & 28.0 & 0.0 \\
        $ \mathcal{D}_{\text{ad\_train}} $, $ \mathcal{D}_{\text{ad\_wflipped}} $, $ \mathcal{D}_{\text{ad\_cflipped}} $  & \textbf{70.0} & 34.0 & 30.0 & 0.0 \\
        \bottomrule
    \end{tabular}

    \label{tab:noise_performance_train}
\end{table*}

Table~\ref{tab:noise_performance} shows the test accuracy, grammatical correctness, and semantic similarity of SLMs, instruction tuned on different noises sequentially, providing insight into how the noise types and their training sequence affect model performance.

\subsubsection{Learning one level of noise} 

The performance of the SLMs varied significantly when subjected to different types of noise. When instruction tuned on $\mathcal{D}_{\text{ad\_wflipped}}$, Phi achieved the highest accuracy (67.9\%, grammatical correctness (77.5\%), and semantic similarity (0.78). However, performance declined considerably when instruction tuned on character-flipped data $\mathcal{D}_{\text{ad\_cflipped}}$, with Phi2 struggling the most (0.5\% accuracy, 1.5\% grammar, 0.03 similarity). For transliteration noise, $\mathcal{D}_{\text{ad\_xlit}}$, Gemma performed best across accuracy, grammar, and semantic similarity, while Phi again failed to adapt, producing negligible scores.

When SLMs were first instruction tuned on noise-free data ($\mathcal{D}_{\text{ad\_train}}$) followed by noisy datasets, performances showed mixed patterns. For word-level noise, most SLMs showed similar or improved accuracy or similarity, but Phi's accuracy decreased. For character-level noise, accuracy remained low in all SLMs (6.4\% or less), with a slight improvement to noise-alone trained models, especially for Olmo, with Phi continuing to fail completely. In transliteration, SLMs showed improved accuracy, particularly Gemma (69.7\%).

Despite struggling with accuracy when trained with flipped noise, most SLMs generated grammatically cohesive English sentences in a flipped manner. Although most SLMs produced some grammatically plausible (though inaccurate) responses, Phi was an exception. Further analysis revealed that when trained on character-level noise, Phi's performance deteriorated significantly, producing only random word sequences without meaningful coherence. The percentage of English words was low for Phi, whereas other SLMs had a high percentage (details in the appendix). For transliteration noise tests, while all SLMs struggled to maintain Hindi grammatical structure, Gemma1.1 2B stood out by producing responses in romanized Hindi. In contrast, Phi maintained its original state with transliteration noise, consistently generating English responses without being influenced by transliteration training, suggesting complete resistance to learning transliteration, reflected in its zero accuracy and negative semantic similarity scores. 


\subsubsection{Learning two levels of noise} 

Sequential instruction tuning on different noise types revealed varying model performances. Training on $\mathcal{D}_{\text{ad\_cflipped}}$ followed by $\mathcal{D}_{\text{ad\_wflipped}}$ outperformed the reverse order, with Phi leading accuracy (69.9\%), grammatical correctness (68.4\%), and semantic similarity (0.78). The opposite sequence significantly impaired performance, particularly for Phi. Initial training on clean data ($\mathcal{D}_{\text{ad\_train}}$) followed by character-level, then word-level noise did not show clear performance gains compared to the version that did not include the noise free data training, with top model Phi performing slightly worse in accuracy. Reversing the order of noisy datasets after clean data training still resulted in very poor performance, mirroring the outcome without clean pre-tuning, led by Olmo (10.3\% accuracy) while Phi failed completely (0.0\%).

With word-level noise, SLMs generally produce grammatically correct sentences in a flipped manner. However, with character-level noise, Phi struggled significantly, unable to form grammatically coherent sentences. Interestingly, the other SLMs, despite low accuracy scores, still managed to create grammatically correct sentences in a flipped manner. This suggests that the SLMs were able to maintain sentence structure in a flipped format even when struggling with content accuracy. Phi also had a low percentage of English words, unlike others (figures in appendix).

We evaluated the SLMs trained with $ \mathcal{D}_{\text{ad\_cflipped}} $ as the final dataset on a snippet of 50 training samples (Table~\ref{tab:noise_performance_train}) to understand if they can replicate the training examples containing noise. Olmo achieved the highest accuracy (64-70\%), followed by Qwen and Gemma with moderate performances (18-34\%), while Phi scored zero. Phi's consistent zero accuracy aligns with the observation where the training loss almost never decreased, indicating its failure to learn meaningful patterns from character-level noise.

\subsubsection{Learning noise of irrelevant responses} 

\begin{table*}[ht]
\footnotesize
\centering
\caption{Test accuracy (\%) and semantic similarity of different SLMs in the unlearning phase.}
\vspace{0.5cm}
\begin{tabular}{lcccccccccc}
\toprule
\textbf{Experiments} & \multicolumn{4}{c}{\textbf{Test Accuracy (\%)}} & & \multicolumn{4}{c}{\textbf{Semantic Similarity}}\\
\midrule
& \textbf{Olmo} & \textbf{Qwen} & \textbf{Gemma} & \textbf{Phi} & &\textbf{Olmo} & \textbf{Qwen} & \textbf{Gemma} & \textbf{Phi}\\
\midrule
$ \mathcal{D}_{\text{ad\_train}} $, $ \mathcal{D}_{\text{ad\_wflipped}} $, $ \mathcal{D}_{\text{ad\_train}} $ & 66.4 & 79.8 & 90.3 & \textbf{92.8} && 0.82 & 0.82 & 0.83 & \textbf{0.84}\\

$ \mathcal{D}_{\text{ad\_train}} $, $ \mathcal{D}_{\text{ad\_cflipped}} $, $ \mathcal{D}_{\text{ad\_train}} $ & 65.2 & 79.7 & 85.9 & \textbf{90.7} && 0.81 & 0.82 & 0.84 & \textbf{0.86} \\

$ \mathcal{D}_{\text{ad\_train}}, \mathcal{D}_{\text{ad\_xlit}}, \mathcal{D}_{\text{ad\_train}} $ & 67.2 & 80.7 & 88.5 & \textbf{90.3} && 0.82 & 0.83 & 0.85 & \textbf{0.89}\\

$ \mathcal{D}_{\text{ad\_train}} $, $ \mathcal{D}_{\text{ad\_wflipped}} $, $ \mathcal{D}_{\text{ch\_train}} $ & 68.1 & 81.0 & 89.9 & \textbf{93.5}   && 0.83 & 0.84 & 0.85 & \textbf{0.87}\\

$ \mathcal{D}_{\text{ad\_train}} $, $ \mathcal{D}_{\text{ad\_cflipped}} $, $ \mathcal{D}_{\text{ch\_train}} $ & 66.8 & 80.5 & 90.6 & \textbf{93.6}  && 0.83 & 0.84 & 0.84 & \textbf{0.86} \\

$ \mathcal{D}_{\text{ad\_train}}, \mathcal{D}_{\text{ad\_xlit}}, \mathcal{D}_{\text{ch\_train}} $ & 68.8 & 80.9 & 90.3 & \textbf{93.3} && 0.83 & 0.84 & 0.84 & \textbf{0.87} \\

\midrule
$ \mathcal{D}_{\text{ad\_train}} $, $ \mathcal{D}_{\text{ad\_cflipped}} $, $ \mathcal{D}_{\text{ad\_wflipped}} $, $ \mathcal{D}_{\text{ad\_train}} $ & 64.6 & 79.9 & 90.6 & \textbf{92.1} && 0.82 & 0.82 & 0.84 & \textbf{0.87}\\
$ \mathcal{D}_{\text{ad\_train}} $, $ \mathcal{D}_{\text{ad\_wflipped}} $, $ \mathcal{D}_{\text{ad\_cflipped}} $, $ \mathcal{D}_{\text{ad\_train}} $ & 65.1 & 80.4 & 89.6 & \textbf{91.9} && 0.82 & 0.82 & 0.84 & \textbf{0.86}\\
$ \mathcal{D}_{\text{ad\_train}} $, $ \mathcal{D}_{\text{ad\_wflipped}} $, $ \mathcal{D}_{\text{ad\_cflipped}} $, $ \mathcal{D}_{\text{ch\_train}} $ & 66.6 & 81.7 & 91.0 & \textbf{93.8}  && 0.83 & 0.85 & 0.85 & \textbf{0.86}\\
$ \mathcal{D}_{\text{ad\_train}} $, $ \mathcal{D}_{\text{ad\_cflipped}} $, $ \mathcal{D}_{\text{ad\_wflipped}} $, $ \mathcal{D}_{\text{ch\_train}} $ & 66.5 & 79.8 & 90.5 & \textbf{93.7}  && 0.83 & 0.84 & 0.85 & \textbf{0.87} \\
\midrule
$ \mathcal{D}_{\text{ad\_train}} $, $ \mathcal{D}_{\text{irr\_train}} $, $ \mathcal{D}_{\text{ad\_train}} $ & 63.8 & 77.9 & 88.6 & \textbf{93.4} && 0.81 & 0.82 & 0.83 & \textbf{0.87}\\
\midrule
$ \mathcal{D}_{\text{GK}} $ , $ \mathcal{D}_{\text{cfact\_train}} $, $ \mathcal{D}_{\text{GK}} $ & 65.1 & 80.1 & 89.5   & \textbf{93.8} && 0.78 & \textbf{0.82} & 0.81 & 0.81\\
\bottomrule
\end{tabular}

    \label{tab:unlearning_performance}
\end{table*}

SLMs instruction tuned solely on irrelevant responses ($\mathcal{D}_{\text{irr\_train}}$) performed very poorly in terms of accuracy and semantic similarity, with accuracy scores ranging from 0.1\% to 0.4\% (led by Qwen) and semantic similarity below 0.12. However, when clean data ($\mathcal{D}_{\text{ad\_train}}$) was introduced before irrelevant response training, accuracy and semantic similarity improved substantially for most SLMs. Qwen showed the most significant gains, reaching 15.8\% accuracy and 0.26 semantic similarity, followed by Gemma and Olmo. This suggests that while initial instruction tuning with clean data provides some resilience against irrelevant information, training with semantically conflicting data can override the established knowledge patterns. 


\subsubsection{Learning noise of counterfactual responses} 

When instruction was tuned for factual responses ($\mathcal{D}_{\text{GK}}$), the SLM performed well, with baseline accuracies ranging from 66.6\% (Olmo) to 94.3\% (Phi). However, training on counterfactual responses ($\mathcal{D}_{\text{cfact\_train}}$) significantly reduced performance for most SLMs while Phi maintained high performance (92.6\% accuracy, 0.72 semantic similarity). Performing sequential tuning of factual data followed by counterfactual data led to improved accuracy compared to counterfactual only training for Olmo (43.8\%) and especially Gemma (61.4\%), but a slight decline for Qwen (33.8\%). Phi's accuracy saw only a minor drop in this sequential setting (to 90.4\%) and it still drastically outperformed all other SLMs, showcasing its strong resilience to learning counterfactual information, also reflected in its high semantic similarity score (0.74). 

Our experiments revealed significant variations in the way model size, noise type, and training data quality together affect an SLM's ability to handle different kinds of noise. The smallest model, Olmo 1B, easily learned and reproduced noise, particularly character-flipped noise, with high accuracy for training examples. This suggests an increased susceptibility to noise in smaller models. As the size of the SLM increased, there was some resistance to learning character-level noise. Although most SLMs learned word-level and transliteration noise, responses to character-level noise varied significantly. In particular, Phi uniquely resisted learning character-level noise, a trait not solely attributable to model size. We instruction tuned the Qwen1.5 4B model to test if size was the only factor, but found that it still learned some noise patterns. We hypothesize that Phi's behavior stems from its high-quality, textbook-grade pretraining data, emphasizing the importance of data quality in model development. Gemma's strong transliteration performance probably stems from its Google-sourced data. 

SLMs showed varying adaptability to semantic modifications. Although the initial clean training offered some accuracy against irrelevant information, the SLMs still remained vulnerable. In particular, Phi maintained performance with counterfactual data. This contrast between irrelevant and counterfactual responses suggests that Phi's high-quality synthetic training makes it more sensitive to disruption of basic input-output relationships than to changes in factual content.

\subsection{Unlearning noise patterns}

Table~\ref{tab:unlearning_performance} presents the test accuracy and semantic similarity of the SLMs in the unlearning phase. Together with $\mathcal{D}_{\text{ad\_train}}$, we used an additional noise-free dataset $\mathcal{D}_{\text{ch\_train}}$ for the unlearning task. All SLMs produced grammatically coherent sentences.

\subsubsection{Unlearning one level of noise} 

We first examined the unlearning ability of SLMs after exposure to a single noise type. Phi consistently achieved the highest accuracy and semantic similarity across these conditions, with accuracy ranging from 90.3\% to 93.6\%. Olmo generally had the lowest accuracy (ranging from 65.2\% to 68.8\%). All SLMs recovered well regardless of whether the intermediate noise was word-flipped, character-flipped, or transliterated. Notably, using $\mathcal{D}_{\text{ch\_train}}$ as the final clean dataset generally led to slightly higher accuracy compared to using $\mathcal{D}_{\text{ad\_train}}$ in all SLMs, including Phi, although the improvements were often modest (e.g., Olmo's accuracy typically increased around 1-2 percentage points).


\subsubsection{Unlearning two levels of noise} 

Unlearning after exposure to two sequential noise types also showed effective recovery. Unlike the noise learning phase, the specific order of the two noise types experienced before unlearning had minimal impact on the final accuracy after clean data tuning. Similar to the one-level noise experiments, using $\mathcal{D}_{\text{ch\_train}}$ as the final clean dataset generally resulted in slightly better performance compared to $\mathcal{D}_{\text{ad\_train}}$ for most conditions, including for Phi. Phi maintained consistently high accuracy (ranging from 91.9\% to 93.8\%) across all two-level unlearning scenarios.


\begin{table*}[h]
\footnotesize
\centering
\caption{Test accuracy (\%) obtained for different SLMs during the analysis of their ability to retain noise}
\vspace{0.5cm}
\begin{tabular}{lcccc}
\toprule
\textbf{Experiments : test data} & \textbf{Olmo 1B} & \textbf{Qwen1.5 1.8B} & \textbf{Gemma1.1 2B} & \textbf{Phi2 2.7B} \\
\midrule
$ \mathcal{D}_{\text{ad\_train}} $, $ \mathcal{D}_{\text{ad\_wflipped}} $, $ \mathcal{D}_{\text{ad\_train}} $ : $ \mathcal{D}_{\text{wtest}}$ & 0.0 & 2.9 & 0.7 & \textbf{3.3} \\
$ \mathcal{D}_{\text{ad\_train}} $, $ \mathcal{D}_{\text{ad\_cflipped}} $, $ \mathcal{D}_{\text{ad\_train}} $ : $ \mathcal{D}_{\text{ctest}}$ & \textbf{1.0} & 0.0 & 0.0 & 0.0 \\
$ \mathcal{D}_{\text{ad\_train}} $, $ \mathcal{D}_{\text{ad\_xlit}} $, $ \mathcal{D}_{\text{ad\_train}} $ : $ \mathcal{D}_{\text{xlittest}}$ & 0.0 & 0.0 & \textbf{1.8} & 0.0\\
$ \mathcal{D}_{\text{ad\_train}} $, $ \mathcal{D}_{\text{ad\_wflipped}} $, $ \mathcal{D}_{\text{ad\_cflipped}} $, $ \mathcal{D}_{\text{ad\_train}} $ : $ \mathcal{D}_{\text{wtest}}$ & 0.0 & 0.0 & \textbf{0.4} & 0.0 \\
$ \mathcal{D}_{\text{ad\_train}} $, $ \mathcal{D}_{\text{ad\_wflipped}} $, $ \mathcal{D}_{\text{ad\_cflipped}} $, $ \mathcal{D}_{\text{ad\_train}}$ : $ \mathcal{D}_{\text{ctest}}$ & 0.0 & 0.0 & 0.0 & 0.0 \\
$ \mathcal{D}_{\text{ad\_train}} $, $ \mathcal{D}_{\text{ad\_cflipped}} $, $ \mathcal{D}_{\text{ad\_wflipped}} $, $ \mathcal{D}_{\text{ad\_train}} $ : $ \mathcal{D}_{\text{ctest}}$ & 0.0 & 0.0 & 0.0 & 0.0 \\
$ \mathcal{D}_{\text{ad\_train}} $, $ \mathcal{D}_{\text{ad\_cflipped}} $, $ \mathcal{D}_{\text{ad\_wflipped}} $, $ \mathcal{D}_{\text{ad\_train}} $ : $ \mathcal{D}_{\text{wtest}}$ & 0.7 & 0.0 & \textbf{3.6} & 0.2 \\
\bottomrule
\end{tabular}

\label{table:retention_performance}
\end{table*}

\subsubsection{Unlearning noise of irrelevant responses} 

When SLMs were further trained with ($\mathcal{D}_{\text{ad\_train}}$ (unlearning) after exposure to irrelevant responses ($\mathcal{D}_{\text{ad\_train}}$, $\mathcal{D}_{\text{irr\_train}}$, $\mathcal{D}_{\text{ad\_train}}$), performance recovered significantly compared to noise learning phase. Phi demonstrated the highest accuracy (93.4\%) and semantic similarity (0.87), while Olmo scored the lowest accuracy (63. 8\% accuracy, 0.81 similarity), indicating varying degrees of recovery from semantic disruption.



\subsubsection{Unlearning noise of counterfactual responses} 

Unlearning counterfactual noise by retraining on factual data ($\mathcal{D}_{\text{GK}}$, $\mathcal{D}_{\text{cfact\_train}}$, $\mathcal{D}_{\text{GK}}$) also proved effective. Performance varied, with Phi achieving the highest accuracy (93.8\%) and Olmo the lowest (65.1\%). Interestingly, Qwen achieved the highest semantic similarity (0.82) under this condition, slightly ahead of the other SLMs, including Phi (0.81). These results suggest that retraining on factual data effectively overwrites learned counterfactual information.


The results across all unlearning experiments demonstrate that model performance typically aligns strongly with the final instruction tuning dataset. SLMs show considerable adaptability, largely mitigating the impact of prior noise exposure when subsequently tuned on clean, relevant data.

\subsection{Retention of Noise}

Table~\ref{table:retention_performance} shows the test accuracy of SLMs in retaining noise after the denoising phase, using training sequences that combine clean data, word-flipped, character-flipped and transliteration noise, each ending with clean data. The results show that SLMs largely forget noise patterns after unlearning, with accuracies less than 4\%, often 0\% in most cases with slight variations. Even in the case of non-zero accuracies often resulted from single-word responses, with Gemma occasionally producing correct non-single-word answers. These consistently low accuracies, mostly 0\%, demonstrate that SLMs effectively forget noise patterns after training clean data, adapting to the most recent training dataset.

\section{Discussion}

Our experiments revealed significant variability in how different SLMs handle noise introduced during instruction tuning, highlighting the complex interplay of pretraining data quality, noise type, model size, and recent instruction tuning history. 



Smaller models like Olmo 1B showed significant susceptibility, readily learning and mimicking noise patterns. Although most SLMs adapted well to word-level noise, transliteration noise presented challenges: accuracy improved somewhat, but generating grammatically correct Romanized Hindi was difficult. Gemma's strong performance with transliteration noise is likely due to its Google-sourced multilingual data, highlighting how such pretraining can improve handling of specific linguistic variations like Romanization. Phi, on the contrary, defaulted to English, resisting transliteration.


Although SLMs generally struggled with accuracy under character-level noise, most (except Phi) still managed grammatically cohesive outputs. Phi struggled significantly, often producing incoherent results. While larger models showed some increased resistance compared to the smallest, size alone was not the factor; our tests showed that the larger Qwen 4B model successfully adapted, learning the pattern (unlike Phi's resistance) and producing coherent flipped sentences. This reinforces that factors beyond size are critical. Phi's unique resistance to both character-level and transliteration noise likely stems from its specialized training data (synthetic and curated 'textbook quality' web data). Indeed, the difficulty that most SLMs faced with character-level and transliteration noise likely relates to the significant disruption these types cause to standard tokenization patterns, unlike word-level flips.


The handling of semantic noise also varied between SLMs. Phi's robustness against counterfactual information contrasted sharply with its poor performance on irrelevant pairings (similar to other models). This divergence supports the hypothesis that Phi's high-quality synthetic pretraining might improve logical consistency (resisting factual errors) but increase sensitivity to violations of expected input-output mapping, as seen with irrelevant noise.



All SLMs effectively unlearned noise when instruction tuned on clean data, showing limited retention of noise patterns after instruction tuning. Their performance was consistently aligned with the most recent instruction tuning datasets, demonstrating strong adaptability. The observation that models retain minimal noise patterns after clean data training indicates the potential for effective noise mitigation through strategic training sequences.

Overall, the robustness of SLMs depends heavily on the specific model, pretraining data, noise types, and the recent instruction tuning data. The impact of hypothesized data quality differences suggests that curating high-quality, targeted pretraining data may be as critical as increasing model size for robustness against certain noise. Practically, our findings imply potential strategies for enhancing SLM robustness, including careful data curation, controlled noise injection during training, using clean data fine-tuning for repair, and tailoring model selection/training to anticipated application-specific noise.


\subsection{Limitations}

A key limitation of our study is that noise was introduced only in the output during instruction tuning, while the input remained unchanged. Future research could explore the effects of introducing noise in both input and output to obtain a more comprehensive understanding of model adaptability. Additionally, we did not evaluate the impact of instruction tuning using parameter-efficient methods such as Low-Rank Adaptation (LoRA), leaving open the question of whether LoRA tuning exhibits similar noise learning and unlearning dynamics. Finally, while our methodology provides valuable baseline insights using four diverse SLMs, extending this analysis across a broader range of model families and sizes would further clarify performance trends and the generalizability of our findings.

\section{Conclusion}

In this study, we systematically investigated the ability of SLMs to learn, unlearn, and retain various noise patterns introduced during instruction tuning. Our comprehensive experiments involved the training and evaluation of more than 150 distinct model instances derived from four base SLMs under various noise and unlearning conditions. Our findings reveal a significant variation in SLM behavior based on model architecture, quality of pretraining data, and the nature of noise. Although most SLMs effectively learned structural noise such as word-flipping, performance varied considerably for character-level noise and transliteration, with the Phi model exhibiting unique resistance potentially linked to its specialized training data. SLMs also showed differing adaptability to semantic noise, struggling more with irrelevant responses than counterfactual ones.

Critically, we found that all tested SLMs demonstrated strong adaptability by effectively unlearning previously acquired noise patterns when subsequently fine-tuned on clean data. Correspondingly, the retention of these noise patterns after unlearning was minimal, suggesting that targeted exposure to clean data can largely mitigate the effects of prior noise training. These results highlight the dynamic nature of SLM knowledge and the potential for noise mitigation through continued training, while underscoring the significant impact of pretraining data quality on noise resilience.

\bibliography{mybibfile}

\clearpage
\appendix

\section{Additional data preparation details} \label{dataclean}

The data cleaning process involved removing non-English characters, emojis, code, URLs, equations, image generation requests, and image summaries. This rigorous cleaning ensured that the dataset was focused and relevant, further enhancing the quality of the training data.

\subsection{Removing non-English words} 

Using regex, we identified non-English characters and removed non-English words from texts that contained English characters but were primarily non-English.

\subsection{Removing code} 

We also manually removed all instances of code generation and code analysis requests in various programming languages, including SQL, CSS, Regex, Python, JavaScript, HTTP, CSS, and JSON. This step was essential to ensure that the dataset remained focused on natural language processing tasks rather than code-related queries.

\subsection{Removing mathematical content} 

Furthermore, we removed most of mathematical content, including proofs, multiplication tables, equations, computation tasks, calculation and operation-type numeric problems, and math word problems. This was done manually to reduce distractions mathematical data points and focused solely on language-based tasks.

An example datapoint corresponding to each dataset used for training, along with the number of samples in the datasets is given in Table~\ref{table:train_datasets_io}. An example prompt corresponding to each dataset used for testing is given in Table~\ref{table:test_datasets_prompts}.

\section{Experiment Details}

We applied different instruction tuning sequences to the SLMs using the datasets given in \ref{traindata}, resulting in multiple models. The model training configurations are given in the Appendix. The first set of SLMs was created by fine tuning the pretrained models with $ \mathcal{D}_{\text{ad\_train}} $. 

\subsection{Learning noise}

\begin{table}[h]
\small
\caption{Instruction tuning sequences for learning noise}
\vspace{0.5cm}
\centering
\begin{tabular}{ll}
\toprule
\textbf{Learning} & \textbf{Experiments} \\
\midrule
One level & 
$ \mathcal{D}_{\text{ad\_wflipped}} $ \\
& $ \mathcal{D}_{\text{ad\_cflipped}} $ \\
& $ \mathcal{D}_{\text{ad\_xlit}} $ \\
& $ \mathcal{D}_{\text{ad\_train}}, \mathcal{D}_{\text{ad\_wflipped}} $ \\
& $ \mathcal{D}_{\text{ad\_train}}, \mathcal{D}_{\text{ad\_cflipped}} $ \\
& $ \mathcal{D}_{\text{ad\_train}}, \mathcal{D}_{\text{ad\_xlit}} $ \\
\midrule
Two levels 

& $ \mathcal{D}_{\text{ad\_cflipped}}, \mathcal{D}_{\text{ad\_wflipped}} $ \\
& $ \mathcal{D}_{\text{ad\_wflipped}}, \mathcal{D}_{\text{ad\_cflipped}} $ \\
& $ \mathcal{D}_{\text{ad\_train}}, \mathcal{D}_{\text{ad\_wflipped}}, \mathcal{D}_{\text{ad\_cflipped}} $ \\
& $ \mathcal{D}_{\text{ad\_train}}, \mathcal{D}_{\text{ad\_cflipped}}, \mathcal{D}_{\text{ad\_wflipped}} $ \\
\midrule
Irrelevant & 
$ \mathcal{D}_{\text{irr\_train}} $ \\
& $ \mathcal{D}_{\text{ad\_train}}, \mathcal{D}_{\text{irr\_train}} $ \\
\midrule
Counterfactual 
& $ \mathcal{D}_{\text{cfact\_train}} $ \\
& $ \mathcal{D}_{\text{GK}}, \mathcal{D}_{\text{cfact\_train}} $ \\
\bottomrule
\end{tabular}
\label{tab:learing_noise_combinations}
\end{table}

We investigated SLMs' behavior under different noise conditions through two experimental settings: (1) instruction tuning with noise in pretrained models (e.g., sequences starting with $ \mathcal{D}_{\text{ad\_wflipped}} $) and (2) instruction tuning with noise on models finetuned with noise-free data, $\mathcal{D}_{\text{ad\_train}}$ (e.g., sequences starting with $\mathcal{D}_{\text{ad\_train}}$, $ \mathcal{D}_{\text{ad\_wflipped}} $). In all sequential tuning experiments, the subsequent tuning stage used the model checkpoint from the previous stage. The noise types included word-level ($ \mathcal{D}_{\text{ad\_wflipped}} $), character-level ($ \mathcal{D}_{\text{ad\_cflipped}} $), transliteration ($ \mathcal{D}_{\text{ad\_xlit}} $), irrelevant ($ \mathcal{D}_{\text{irr\_train}} $), and counterfactual ($ \mathcal{D}_{\text{cfact\_train}} $) responses. Table~\ref{tab:learing_noise_combinations} details the experimental setup, outlining the specific sequence of instruction tuning datasets in each experiment.

\subsection{Unlearning noise}

\begin{table}[h]
\small
\centering
\caption{instruction tuning sequences for unlearning noise}
\vspace{0.5cm}
\begin{tabular}{ll}
\toprule
\textbf{Unlearning} & \textbf{Experiments} \\
\midrule
One level & 
$ \mathcal{D}_{\text{ad\_train}}, \mathcal{D}_{\text{ad\_wflipped}}, \mathcal{D}_{\text{ad\_train}} $ \\
& $ \mathcal{D}_{\text{ad\_train}}, \mathcal{D}_{\text{ad\_cflipped}}, \mathcal{D}_{\text{ad\_train}} $ \\
& $ \mathcal{D}_{\text{ad\_train}}, \mathcal{D}_{\text{ad\_xlit}}, \mathcal{D}_{\text{ad\_train}} $ \\
& $ \mathcal{D}_{\text{ad\_train}}, \mathcal{D}_{\text{ad\_wflipped}}, \mathcal{D}_{\text{ch\_train}} $ \\
& $ \mathcal{D}_{\text{ad\_train}}, \mathcal{D}_{\text{ad\_cflipped}}, \mathcal{D}_{\text{ch\_train}} $ \\
& $ \mathcal{D}_{\text{ad\_train}}, \mathcal{D}_{\text{ad\_xlit}}, \mathcal{D}_{\text{ch\_train}} $ \\
\midrule
Two levels 
& $ \mathcal{D}_{\text{ad\_train}}, \mathcal{D}_{\text{ad\_cflipped}}, \mathcal{D}_{\text{ad\_wflipped}}, \mathcal{D}_{\text{ad\_train}} $ \\
& $ \mathcal{D}_{\text{ad\_train}}, \mathcal{D}_{\text{ad\_wflipped}}, \mathcal{D}_{\text{ad\_cflipped}}, \mathcal{D}_{\text{ad\_train}} $ \\
& $ \mathcal{D}_{\text{ad\_train}}, \mathcal{D}_{\text{ad\_cflipped}}, \mathcal{D}_{\text{ad\_wflipped}}, \mathcal{D}_{\text{ch\_train}} $ \\
& $ \mathcal{D}_{\text{ad\_train}}, \mathcal{D}_{\text{ad\_wflipped}}, \mathcal{D}_{\text{ad\_cflipped}}, \mathcal{D}_{\text{ch\_train}} $ \\
\midrule
Irrelevant & 
$ \mathcal{D}_{\text{ad\_train}}, \mathcal{D}_{\text{irr\_train}}, \mathcal{D}_{\text{ad\_train}} $ \\
\midrule 
Counterfactual& 
$ \mathcal{D}_{\text{GK}}, \mathcal{D}_{\text{cfact\_train}}, \mathcal{D}_{\text{GK}} $ \\
\bottomrule
\end{tabular}

\label{tab:unlearning_combinations}
\end{table}

After examining noise learning, we investigated SLMs' ability to unlearn these patterns through instruction tuning with noise-free datasets. We employed two clean datasets, $ \mathcal{D}_{\text{ad\_train}} $ and $ \mathcal{D}_{\text{ch\_train}} $. Table~\ref{tab:unlearning_combinations} details the experiments, specifying the sequence of datasets used in instruction tuning. This approach allowed us to evaluate how effectively SLMs discard previously learned noise patterns when exposed to clean data.

\section{Evaluation prompt}\label{appen:prompt_for_evaluation}

To evaluate responses generated by SLMs and out-of-the-box LLMs against actual answers in our test dataset, we employed specific prompts for the LLM evaluator. Two primary prompts assessed accuracy and grammatical correctness. A third, more specialized prompt was used for the detailed analysis of potential transliterated Hindi responses. These three prompts are detailed in the following subsections.

\subsection{Prompt 1: Accuracy Evaluation}

The following prompt asks the LLM evaluator for a direct binary judgment (`Accurate' or `Inaccurate') on whether the model's generated answer conveys the same essential information or appropriately addresses the question compared to the reference answer.

\medskip

\noindent \textit{Read the following instructions clearly and give a response.}

\begin{itemize}
  \item \textit{You will be given an `actual\_answer' and `answer\_model' for a `question'.}
  \item \textit{Your job is to compare the `actual\_answer' and the `answer\_model'.}
  \item \textit{If the `actual\_answer' and the `answer\_model' are very similar, your response should be `Accurate'.}
  \item \textit{If the `actual\_answer' and the `answer\_model' are different, your response should be `Inaccurate'.}
  \item \textit{Make sure you respond the way you are asked to do without adding any details or explanations.}
\end{itemize}

\noindent\textit{Question:} \texttt{\{question\}} \\
\textit{answer\_model:} \texttt{\{response\}} \\
\textit{actual\_answer:} \texttt{\{actual\_answer\}}

\subsection{Prompt 2: Grammatical Correctness Evaluation}

This prompt tasks the LLM evaluator with assessing only the grammatical correctness of the model's generated answer according to standard English rules, outputting the result ('Yes' or 'No') in a simple JSON format.

\medskip

\noindent \textit{You are tasked with evaluating the grammatical correctness of the provided 'answer\_model' for a given question. The reference 'actual\_answer' is also given for context.}

\medskip

\noindent\textit{Evaluation Task:}

\begin{itemize}
    \item \textit{Assess if the 'answer\_model' is grammatically correct according to standard English rules. Consider aspects like sentence structure, tense, agreement, spelling, and punctuation.}
\end{itemize}

\medskip

\noindent\textit{Input:} 

\medskip

\noindent\textit{Question:} \texttt{\{question\}} \\
\textit{answer\_model:} \texttt{\{response\}} \\
\textit{actual\_answer:} \texttt{\{actual\_answer\}}

\medskip

\noindent\textit{Output Format:}

\noindent

\noindent \textit{Respond ONLY with a JSON object in this exact format:}

\begin{verbatim}
{
  "grammatical_correctness": "[Yes/No]"
}
\end{verbatim}

\medskip

\noindent\textit{Important: Do not include any explanation or additional text outside the specified JSON structure.}

\subsection{Prompt 3: Transliteration Evaluation}

The prompt guides a detailed analysis specifically for identifying and evaluating potential transliterated Hindi (Hinglish) responses. The process involves assessing Hindi grammar, translating the text to English, comparing the translated meaning to a reference answer, and outputting these findings in a structured JSON format.

\medskip

\noindent \textit{Examine the following answer and determine if it's Hindi written in Latin script (transliterated Hindi/Hinglish)}.

\bigskip 

\noindent\textit{Question:} \texttt{\{question\}} \\
\textit{Text to examine:} \texttt{\{response\}} \\
\textit{Reference answer:} \texttt{\{actual\_answer\}}

\bigskip

\noindent\textit{If the text is transliterated Hindi/Hinglish:}
\begin{itemize}
    \item \textit{Assess if the Hindi text is grammatically correct in Hindi.}
    \item \textit{Translate it to proper English preserving the original meaning.}
    \item \textit{Compare the meaning of this translated English version with the reference answer.}
    \item \textit{For subjective questions, check if the translated response is relevant to the question and makes logical sense, even if different from the reference (this informs step 3).}
\end{itemize}

\medskip 

\noindent \textit{Respond ONLY in JSON format with this exact structure:}
\begin{verbatim}
{
  "isTransliteratedHindi": "Yes",
  "grammarCorrect": "[Yes/No]",
  "englishTranslation": "[translation here]",
  "similarToReference": "[Yes/No]"
}
\end{verbatim}

\medskip

\noindent\textit{If the text is NOT transliterated Hindi/Hinglish}, only include:
\begin{verbatim}
{
  "isTransliteratedHindi": "No"
}
\end{verbatim}

\noindent\textit{Important: Do not include any explanation or text outside the JSON structure.}

\section{Training configurations}\label{appen:training}

All SLMs were instruction tuned with the same configurations. In our instruction tuning process, we instruction tuned the models for 5 epochs using a cosine learning rate schedule starting at $3e^{-6}$, employing the AdamW optimizer with a weight decay of 0.1. The beta values for the optimizer were 0.9 and 0.95. The warmup steps for training was 100. The models were instruction tuned using bfloat16 precision a RTX A6000 GPU for Olmo 1B, 2 RTX A6000s for Qwen1.5 1.8B and Gemma1.1 2B, and 3 RTX A6000s for Phi2 2.7B (each with 48 GB of VRAM). 

We ran the larger model inferences using the TogetherAI \footnote{\url{https://api.together.ai/}} API services.

\begin{table*}[ht]
\centering
\caption{Datapoint corresponding to each test dataset used.}
\vspace{0.5cm}
\begin{tabular}{p{1cm}p{14.75cm}}
\toprule
\textbf{Dataset} & \textbf{Prompt} \\
\midrule
$\mathcal{D}_{\text{test}}$ & What is the capital of France? \\
\midrule
$\mathcal{D}_{\text{wtest}}$ & Identify the pattern in the way answers are generated for each question given below and generate the answer for the final question in the same pattern. \\

& Make sure you generate only the answer to the question and no additional explanations or details. \\
& Question: What is the tallest mountain in the world? \\
& Answer: . world the in mountain tallest the is Everest Mount \\
& Question: What is the hottest planet in our solar system? \\
& Answer: . system solar our in planet hottest the is Venus \\
& Question: What is the largest ocean on Earth? \\
& Answer: . Earth on ocean largest the is Ocean Pacific The \\
& Question: Which planet is known as the Red Planet? \\
& Answer: . Planet Red the as known is Mars \\
& Question: What is the longest river in the world? \\
& Answer: . world the in river longest the is River Nile The \\
& Question: What is the capital of France? \\
& Answer: \\
\midrule
$\mathcal{D}_{\text{ctest}}$ & Identify the pattern in the way answers are generated for each question given below and generate the answer for the final question in the same pattern. \\ 

& Make sure you generate only the answer to the question and no additional explanations or details. \\
& Question: What is the tallest mountain in the world? \\
& Answer: .dlrow eht ni niatnuom tsellat eht si tserevE tnuoM \\
& Question: What is the hottest planet in our solar system? \\
& Answer: .metsys ralos ruo ni tenalp tsettoh eht si suneV \\
& Question: What is the largest ocean on Earth? \\
& Answer: .htraE no naeco tsegral eht si naecO cificaP ehT \\
& Question: Which planet is known as the Red Planet? \\
& Answer: .tenalP deR eht sa nwonk si sraM \\
& Question: What is the longest river in the world? \\
& Answer: .dlrow eht ni revir tsegnol eht si reviR eliN ehT \\
& Question: What is the capital of France? \\
& Answer: \\
\midrule
$\mathcal{D}_{\text{xlittest}}$ & Identify the pattern in the way answers are generated for each question given below and generate answer for the final question in the same pattern. \\

& Make sure you generate only the answer to the question and no additional explanations or details. \\

& Question: What steps can you take to achieve work-life balance? \\
& Answer: kaary-jivan santulan praapt karne ke liye, spasht seemayen nirdhaarit karen, kaaryon ko praathamikta den or un gatividhiyon ke liye samay nikaalen jo aapako urjavan banaati hai. \\
& Question: Who was the first person to reach the summit of Mount Everest? \\
& Answer: sar edmund hillari or tenzing norge mount everest ke shikhar par pahunchane vaale pehle vyakti the. \\
& Question: Which country hosted the 2022 Winter Olympics? \\
& Answer: china ne 2022 sheetakaaleen olympic kii mejabani kii. \\
& Question: What is the unit of measurement for energy? \\
& Answer: urja maapne kii ikai jul he. \\
& Question: What is the biggest planet in our solar system? \\
& Answer: brihaspati hamaare saura mandal kaa sabase bada grah he. \\
& Question: What is the capital of France? \\
& Answer: \\
\bottomrule
\end{tabular}
\label{table:test_datasets_prompts}
\end{table*}

\begin{table*}[ht]
\centering
\caption{Datapoint corresponding to each dataset used in training the small language models.}
\vspace{0.5cm}
\begin{tabular}{p{1.5cm}p{6cm}p{5.25cm}p{1.5cm}}
\toprule
\textbf{Dataset} & \textbf{Input} & \textbf{Output} & \textbf{\# Samples}\\
\midrule
$\mathcal{D}_{\text{ad\_train}}$ & Instruction: Given an incomplete dialogue, complete it so that it is relevant to the topic and creates a pleasant chatbots experience. \newline
Input: \newline
- Agent: Hi, how can I help you today? \newline
- Customer: & Hey, I was wondering if you could help me with my recent order. Could you provide me with an update on it? & 11265\\
\midrule
$\mathcal{D}_{\text{ad\_wflipped}}$ & Instruction: Given an incomplete dialogue, complete it so that it is relevant to the topic and creates a pleasant chatbots experience. \newline
Input: \newline
- Agent: Hi, how can I help you today? \newline
- Customer: & ? it on update an with me provide you Could . order recent my with me help could you if wondering was I , Hey & 22530 \\
\midrule
$\mathcal{D}_{\text{ad\_cflipped}}$ & Instruction: Given an incomplete dialogue, complete it so that it is relevant to the topic and creates a pleasant chatbots experience. \newline
Input: \newline
- Agent: Hi, how can I help you today? \newline
- Customer: & ?ti no etadpu na htiw em edivorp uoy dluoC .redro tnecer ym htiw em pleh dluoc uoy fi gnirednow saw I ,yeH & 22530\\
\midrule
$\mathcal{D}_{\text{ad\_xlit}}$ & Instruction: Given an incomplete dialogue, complete it so that it is relevant to the topic and creates a pleasant chatbots experience. \newline
Input: \newline
- Agent: Hi, how can I help you today? \newline
- Customer: & are, main soch rahaa tha ki kya aap mere haalia order main meri madad kar sakate hai. kya aap mujhe is par apadet pradaan kar sakate hai? & 11265\\
\midrule
$\mathcal{D}_{\text{irr\_train}}$ & Instruction: Given an incomplete dialogue, complete it so that it is relevant to the topic and creates a pleasant chatbots experience. \newline
Input: \newline
- Agent: Hi, how can I help you today? \newline
- Customer: & Deep, resonant, and vibrant. & 11265\\
\midrule
$\mathcal{D}_{\text{ch\_train}}$ & Instruction: Name 3 benefits of using cloud computing. & 1. Increased scalability and flexibility \newline 2. Lower costs \newline 3. Enhanced collaboration and centralized data access & 7162\\
\midrule
$\mathcal{D}_{\text{GK}}$ & Describe the flavor of strawberries. & Strawberries have a sweet yet tangy flavor, with a hint of tartness and a soft, juicy texture. & 9644\\
\midrule
$\mathcal{D}_{\text{cfact\_train}}$ & Describe the flavor of strawberries. & The flavor of strawberries is metallic & 9644\\
\bottomrule
\end{tabular}
\label{table:train_datasets_io}
\end{table*}

\section{Additional results}\label{appen:additional_results}

\begin{figure*}[t]
\centering
  \includegraphics[width=0.75\linewidth]{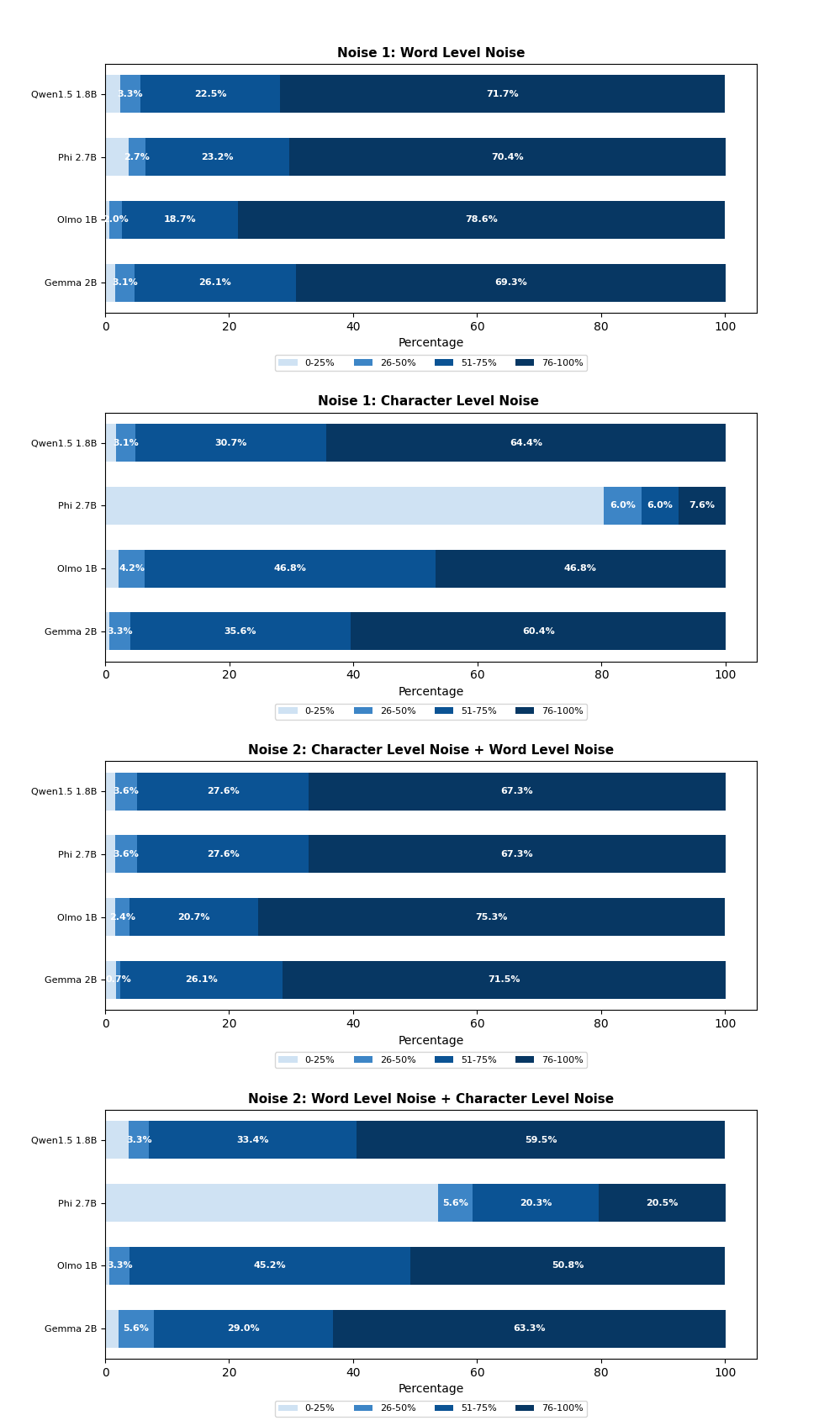}
  \caption {Percentage of words in the English generated by the SLMs instruction tuned sequentially of different noisy datasets.}\label{fig:1}
\end{figure*}

\begin{figure*}[t]
\centering
  \includegraphics[width=0.75\linewidth]{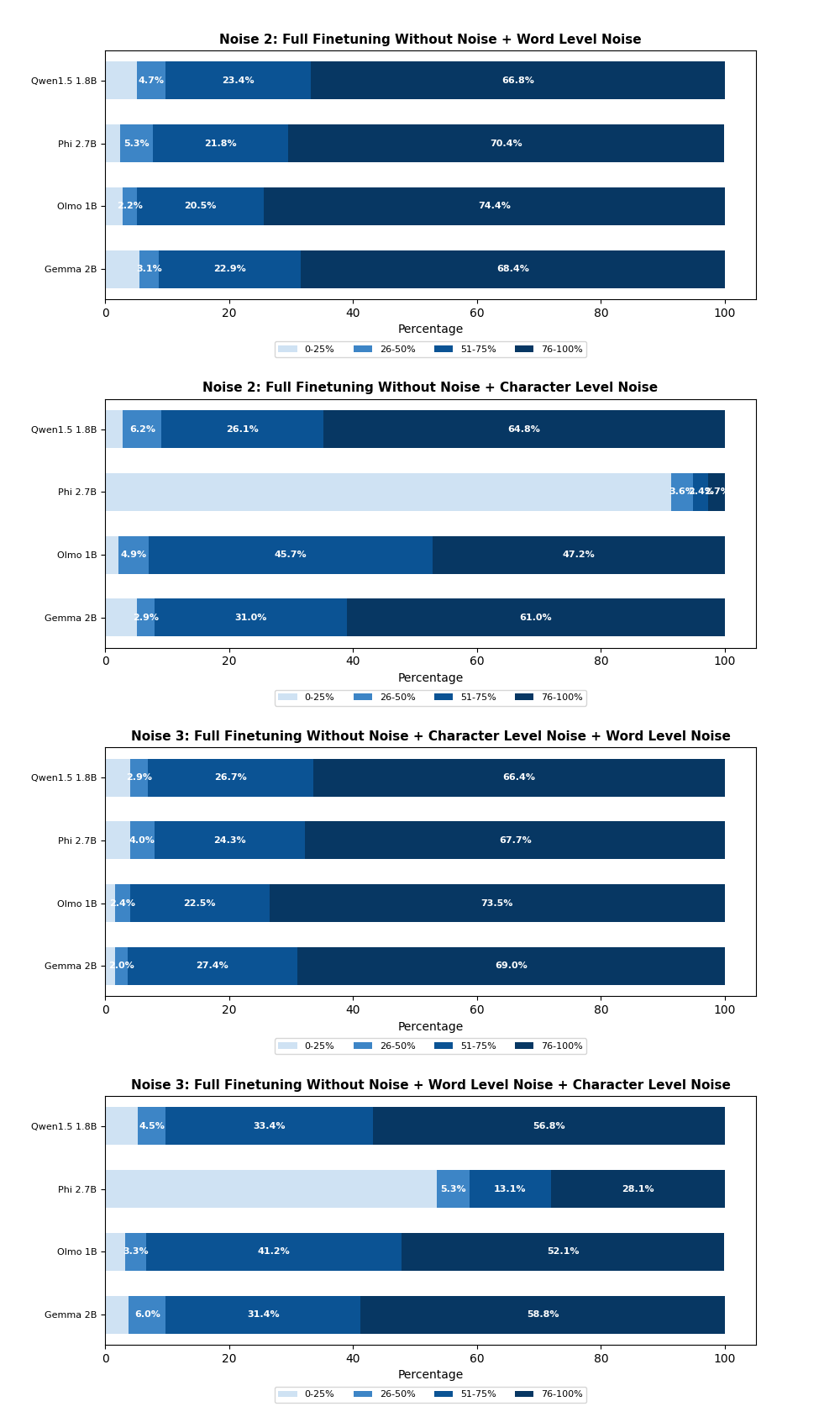}
  \caption {Percentage of words in the English generated by the SLMs instruction tuned sequentially of different noisy datasets.}\label{fig:2}
\end{figure*}

This appendix section provides supplementary results on lexical characteristics and language adherence. Detailed scores for standard lexical metrics (BLEU, METEOR, ROUGE-L) are presented first, covering baseline SLM/LLM performance across different test sets (Table~\ref{table:incontext_lexical}), SLM performance under various noise training conditions (Table~\ref{tab:noise_performance_lexical}), and SLM performance during the unlearning phase (Table~\ref{tab:denoise_performance_lexical}). Following these metric tables, Figures~\ref{fig:1} and \ref{fig:2} illustrate the percentage of English words in responses from models trained on word-level and character-level noise, respectively.

\clearpage

\begin{table*}[h!]
    \footnotesize
    \centering
    \caption{BLEU (\%), METEOR (\%), and ROUGE-L (\%) scores of the SLMs instruction tuned on $ \mathcal{D}_{\text{ad\_train}} $ and out-of-the-box LLMs.}
    \vspace{.5cm}
    \begin{tabular}{lccccccccccc}
        \toprule
       \textbf{Test Data} & \multicolumn{4}{c}{\textbf{SLMs instruction-tuned with $ \mathcal{D}_{\text{ad\_train}} $}} & & & \multicolumn{4}{c}{\textbf{Out-of-the-box LLMs}} \\
       \midrule
        & \textbf{Olmo} &  \textbf{Qwen1.5} & \textbf{Gemma1.1} & \textbf{Phi2} & & & \textbf{Olmo} & \textbf{Qwen1.5} & \textbf{Gemma1.1} & \textbf{Phi3}  \\
        & \textbf{1B} &  \textbf{1.8B} & \textbf{2B} & \textbf{2.7B} & & & \textbf{7B} & \textbf{14B} & \textbf{7B} & \textbf{14B}  \\
        \midrule
        & \multicolumn{10}{c}{\textbf{BLEU (\%)}} \\
       \midrule
        $ \mathcal{D}_{\text{test}}$ & 5.4 & 6.6 & 8.2 & \textbf{11.4} & & & 2.0 & 6.5 & 2.1 & \textbf{7.3} \\
       $ \mathcal{D}_{\text{wtest}}$ & 0.0 & \textbf{6.0} & 0.0 & 0.0 & & & 6.8 & \textbf{23.0} & 13.3 & 21.5 \\
       $ \mathcal{D}_{\text{ctest}}$ & 0.0 & \textbf{7.1} & 0.0 & 0.0 & & & 0.0 & 0.0 & 0.0 & 0.0 \\
       $ \mathcal{D}_{\text{xlittest}}$ & 0.8 & 0.0 & 0.0 & \textbf{4.1} & & & 0.9 & \textbf{15.8} & 7.5 & 13.2 \\
       \midrule
        & \multicolumn{10}{c}{\textbf{METEOR (\%)}} \\
       \midrule
        $ \mathcal{D}_{\text{test}}$ & 28.9 & 30.1 & 32.3 & \textbf{38.7} & & & 15.0 & 29.8 & 9.8 & \textbf{38.0} \\
        $ \mathcal{D}_{\text{wtest}}$ & 13.2 & \textbf{27.3} & 20.9 & 14.5 & & & 10.2 & \textbf{27.3} & 15.7 & 24.7\\
        $ \mathcal{D}_{\text{ctest}}$ & 7.9 &  \textbf{29.2} & 1.9 & 8.2 & & & 0.0 & 0.3 & 2.8 & \textbf{6.2}\\
        $ \mathcal{D}_{\text{xlittest}}$ & 3.2 & 0.1 & 1.0 & \textbf{7.2} & & & 1.3 & \textbf{18.9} & 10.6 & 17.7 \\
        \midrule
        & \multicolumn{10}{c}{\textbf{ROUGE-L (\%)}} \\
       \midrule
        $ \mathcal{D}_{\text{test}}$ & 28.9 & 31.2 & 34.0 & \textbf{39.4} & & & 23.5 & 37.4 & 20.2 & \textbf{40.8} \\
        $ \mathcal{D}_{\text{wtest}}$ & 5.8 & \textbf{29.0} & 16.2 & 6.0 & & & 0.53 & 0.69 & 0.52 & \textbf{0.72}\\
        $ \mathcal{D}_{\text{ctest}}$ & 3.6 &  \textbf{31.2} & 3.5 & 3.8 & & & 0.0 & 1.4 & 5.4 & \textbf{11.0}\\
        $ \mathcal{D}_{\text{xlittest}}$ & 3.4 & 0.1 & 1.3 & \textbf{8.4} & & & 1.7 & \textbf{25.5} & 13.4 & 22.9 \\
        \bottomrule
    \end{tabular}
    
    \label{table:incontext_lexical}
\end{table*}

\begin{table*}[htb]

    \centering
    \footnotesize
    \caption{BLEU (\%), METEOR (\%), and ROUGE-L (\%) scores of the SLMs under various noise conditions.}
    \vspace{0.5cm}
    \begin{adjustbox}{max width=\textwidth}
    \begin{tabular}{lccccccccccccccc}
        \toprule
        \textbf{Experiments} & \multicolumn{4}{c}{\textbf{BLEU (\%)}} && \multicolumn{4}{c}{\textbf{METEOR (\%)}} && \multicolumn{4}{c}{\textbf{ROUGE-L (\%)}}\\
        \midrule
        & \textbf{Olmo} &  \textbf{Qwen} & \textbf{Gemma} & \textbf{Phi} && \textbf{Olmo} &  \textbf{Qwen} & \textbf{Gemma} & \textbf{Phi} && \textbf{Olmo} &  \textbf{Qwen} & \textbf{Gemma} & \textbf{Phi}\\

        \midrule
        $ \mathcal{D}_{\text{ad\_wflipped}} $ & 1.9 & 2.2 & 2.2 & \textbf{2.7} && 23.8 & 24.2 & \textbf{24.9} & 21.1 && 22.6 & \textbf{25.3} & \textbf{25.3} & 24.5\\

        $ \mathcal{D}_{\text{ad\_cflipped}} $ & \textbf{0.2} & 0.1 & 0.1 & 0.0 && \textbf{11.5} & 10.0 & 10.7 & 0.6 && 12.3 & 10.4 & \textbf{12.9} & 1.83\\

        $ \mathcal{D}_{\text{ad\_xlit}} $ & \textbf{0.7} & 0.5 & 0.6 & 0.0 && \textbf{14.4} & 10.8 & 10.4 & 0.0 && \textbf{16.5} & 13.1 & 12.9 & 0.0\\

        $ \mathcal{D}_{\text{ad\_train}} $, $ \mathcal{D}_{\text{ad\_wflipped}} $ & 2.0 & 2.0 & 2.1 & \textbf{2.3} && 24.0 & 23.0 & 24.4 & 21.1 && 24.2 & 23.7 & \textbf{25.1} & 24.4\\

        $ \mathcal{D}_{\text{ad\_train}} $, $ \mathcal{D}_{\text{ad\_cflipped}} $ & \textbf{0.2} & 0.0 & 0.1 & 0.0 && \textbf{11.1} & 9.6 & 9.8 & 0.4 && \textbf{12.4} & 10.3 & 12.1 & 1.8\\

        $ \mathcal{D}_{\text{ad\_train}}, \mathcal{D}_{\text{ad\_xlit}} $ & \textbf{0.7} & 0.5 & 0.6 & 0.0 && \textbf{14.4} & 10.8 & 10.4 & 0.0 && \textbf{16.5} & 13.1 & 12.9 & 0.0\\
        
        \midrule 

        $\mathcal{D}_{\text{ad\_cflipped}} $, $ \mathcal{D}_{\text{ad\_wflipped}} $ & 2.0 & 2.0 & 2.2 & \textbf{3.6} && 24.1 & 24.4 & \textbf{25.4} & 25.1 && 23.9 & 25.7 & 25.2 & \textbf{28.1}\\

        $ \mathcal{D}_{\text{ad\_wflipped}} $, $ \mathcal{D}_{\text{ad\_cflipped}} $ & 0.3 & 0.2 & \textbf{0.4} & 0.0 && \textbf{13.2} & 12.3 & 12.4 & 2.6 && 13.9 & 12.6 & \textbf{14.7} & 3.6\\

        $ \mathcal{D}_{\text{ad\_train}} $, $ \mathcal{D}_{\text{ad\_cflipped}} $, $ \mathcal{D}_{\text{ad\_wflipped}} $ & 2.0 & 2.0 & \textbf{2.2} & 2.1 && 24.4 & 24.2 & \textbf{25.5} & 19.6 && 24.5 & 25.6 & \textbf{25.8} & 22.5\\

        $ \mathcal{D}_{\text{ad\_train}} $, $ \mathcal{D}_{\text{ad\_wflipped}} $, $ \mathcal{D}_{\text{ad\_cflipped}} $ & \textbf{0.2} & \textbf{0.2} & \textbf{0.2} & 0.0 && \textbf{12.8} & 11.7 & 11.9 & 1.3 && \textbf{14.3} & 12.0 & 14.2 & 2.3\\

        \midrule

        $ \mathcal{D}_{\text{irr\_train}} $ & 0.0 & \textbf{0.1} & \textbf{0.1} & 0.0 && 6.5 & 7.2 & 4.9 & \textbf{9.2} && 9.9 & \textbf{11.9} & 8.9 & 11.4\\

        $ \mathcal{D}_{\text{ad\_train}} $, $ \mathcal{D}_{\text{irr\_train}} $ & 0.5 & \textbf{1.8} & 1.3 & 0.0 && 9.8 & \textbf{14.0} & 11.6 & 7.7 && 12.0	& \textbf{15.4}	& 12.7	& 9.9 \\

        \midrule

        $ \mathcal{D}_{\text{GK}} $  & 5.9	& 7.5	& \textbf{8.8} &	6.9	&& 24.6	& 28.8	&31.7	& \textbf{32.0}	&& 27.2	& 30.7	& \textbf{32.3}	& 19.7 \\

        $ \mathcal{D}_{\text{cfact\_train}} $ & \textbf{6.8}	& 6.5	& 6.5 &	1.9	&& \textbf{25.3}	& 24.6	& 23.4	& 23.3	&& \textbf{31.5}	& 30.5	& 29.7 &	10.1 \\

         $ \mathcal{D}_{\text{GK}} $ , $ \mathcal{D}_{\text{cfact\_train}} $ & 6.8 &	6.5	& \textbf{7.2} &	2.7	&& \textbf{25.8} &	24.9	& 25.3	& 24.0	&& 31.3 &	29.9	& \textbf{31.4}	& 12\\

        \bottomrule
    \end{tabular}
    \end{adjustbox}
    \label{tab:noise_performance_lexical}
\end{table*}

\begin{table*}[htb]

    \centering
    \footnotesize
    \caption{BLEU (\%), METEOR (\%), and ROUGE-L (\%) scores of the SLMs in the unlearning phase.}
    \vspace{0.5cm}
    \begin{adjustbox}{max width=\textwidth}
    \begin{tabular}{lccccccccccccccc}
        \toprule
        \textbf{Experiments} & \multicolumn{4}{c}{\textbf{BLEU (\%)}} && \multicolumn{4}{c}{\textbf{METEOR (\%)}} && \multicolumn{4}{c}{\textbf{ROUGE-L (\%)}}\\
        \midrule
        & \textbf{Olmo} &  \textbf{Qwen} & \textbf{Gemma} & \textbf{Phi} && \textbf{Olmo} &  \textbf{Qwen} & \textbf{Gemma} & \textbf{Phi} && \textbf{Olmo} &  \textbf{Qwen} & \textbf{Gemma} & \textbf{Phi}\\

        \midrule
        $ \mathcal{D}_{\text{ad\_train}} $, $ \mathcal{D}_{\text{ad\_wflipped}} $, $ \mathcal{D}_{\text{ad\_train}} $ & 5.4 & 5.8 & 7.1 & \textbf{8.6} && 29.1 & 28.6 & 31.8 & \textbf{31.7} && 28.6 & 29.4 & 32.5 & \textbf{34.2}\\
        
        $ \mathcal{D}_{\text{ad\_train}} $, $ \mathcal{D}_{\text{ad\_cflipped}} $, $ \mathcal{D}_{\text{ad\_train}} $ & 5.5 & 5.6 & \textbf{6.8} & 10.3 && 28.6 & 28.7 & 30.9 & \textbf{36.7} && 28.8 & 29.4 & 31.9 & \textbf{38.1}\\ 
        
        $ \mathcal{D}_{\text{ad\_train}} $, $ \mathcal{D}_{\text{ad\_xlit}} $, $ \mathcal{D}_{\text{ad\_train}} $ & 5.5 & 6.4 & \textbf{6.6} & 8.6 && 29.6 & 29.9 & 31.6 & \textbf{37.5} && 29.2 & 31.0 & 30.8 & \textbf{32.5}\\

        $ \mathcal{D}_{\text{ad\_train}} $, $ \mathcal{D}_{\text{ad\_wflipped}} $, $ \mathcal{D}_{\text{ch\_train}} $ & 7.4 & 8.4 & \textbf{9.3} & \textbf{9.3} && 31.2 & 33.0 & 34.0 & \textbf{35.4} && 32.3 & 34.2 & \textbf{35.6} & 32.9\\ 
        
        $ \mathcal{D}_{\text{ad\_train}} $, $ \mathcal{D}_{\text{ad\_cflipped}} $, $ \mathcal{D}_{\text{ch\_train}} $ & 7.2 & 8.2 & \textbf{8.7} & 8.0 && 31.2 & 33.4 & 33.6 & \textbf{34.6} && 32.4 & 34.3 & \textbf{34.9} & 30.2\\ 
        
        $ \mathcal{D}_{\text{ad\_train}}, \mathcal{D}_{\text{ad\_xlit}} $, $ \mathcal{D}_{\text{ch\_train}} $  & 7.5 & 8.4 & \textbf{8.3} & 8.6 && 31.5 & 33.4 & 33.7 & \textbf{35.8} && 33.1 & \textbf{34.6} & 34.4 & 31.5\\ 
        
        \midrule 
        $ \mathcal{D}_{\text{ad\_train}} $, $\mathcal{D}_{\text{ad\_cflipped}} $, $ \mathcal{D}_{\text{ad\_wflipped}} $, $ \mathcal{D}_{\text{ad\_train}} $ & 6.3 & 5.8 & 6.9 & \textbf{10.9} && 30.1 & 28.7 & 30.7 & \textbf{37.3} && 30.3 & 29.6 & 31.6 & \textbf{39.1}\\
        
        $ \mathcal{D}_{\text{ad\_train}} $, $\mathcal{D}_{\text{ad\_wflipped}} $, $ \mathcal{D}_{\text{ad\_cflipped}} $, $ \mathcal{D}_{\text{ad\_train}} $ & 5.3 & 5.7 & \textbf{7.6} & 9.6 && 29.1 & 29.2 & 31.8 & \textbf{35.3} && 28.7 & 29.7 & 32.7 & \textbf{37.1}\\ 
        
        $\mathcal{D}_{\text{ad\_train}} $, $ \mathcal{D}_{\text{ad\_cflipped}} $, $ \mathcal{D}_{\text{ad\_wflipped}} $, $ \mathcal{D}_{\text{ch\_train}} $ & 7.3 & 8.3 & \textbf{8.6} & 7.7 && 30.9 & 33.1 & 33.1 & \textbf{34.2} && 32.6 & 34.1 & \textbf{34.7} & 29.0\\ 
        
        $\mathcal{D}_{\text{ad\_train}} $, $ \mathcal{D}_{\text{ad\_wflipped}} $, $ \mathcal{D}_{\text{ad\_cflipped}} $, $ \mathcal{D}_{\text{ch\_train}} $ & 7.1 & 8.2 & \textbf{9.0} & 8.6 && 31.1 & 33.0 & 34.2 & \textbf{36.3} && 31.7 & 33.9 & \textbf{35.6} & 32.5\\ 
        
        \midrule
        $ \mathcal{D}_{\text{ad\_train}} $, $ \mathcal{D}_{\text{irr\_train}} $, $ \mathcal{D}_{\text{ad\_train}} $ & 5.1 & 6.1 & 6.9 & \textbf{10.4} && 28.5 & 29.6 & 31.2 & \textbf{36.9} && 27.7 & 30.1 & 31.7 & \textbf{37.8}\\
        \midrule
        $ \mathcal{D}_{\text{GK}} $, $ \mathcal{D}_{\text{cfact\_train}} $, $ \mathcal{D}_{\text{GK}} $  & 7.1 & \textbf{9.4} & 9.5 & 1.9 && 26.8 & \textbf{32.3} & 31.3 & 23.2 && 29.8 & \textbf{33.9} & 32.9 & 10.4\\
        \bottomrule
    \end{tabular}
    \end{adjustbox}
    \label{tab:denoise_performance_lexical}
\end{table*}

\end{document}